\documentclass[lettersize,journal]{IEEEtran}
\usepackage{amsmath,amsfonts}
\usepackage{algorithmic}
\usepackage{algorithm}
\usepackage{array}
\usepackage[caption=false,font=normalsize,labelfont=sf,textfont=sf]{subfig}
\usepackage{textcomp}
\usepackage{stfloats}
\usepackage{url}
\usepackage{verbatim}
\usepackage{graphicx}
\usepackage{cite}
\newcommand{\ba}{\mathbf{a}}\newcommand{\bA}{\mathbf{A}}

\newcommand{\bF}{\mathbf{F}} %
\newcommand{\bg}{\mathbf{g}}
\newcommand{\bh}{\mathbf{h}}

\newcommand{\bK}{\mathbf{K}}

\newcommand{\bO}{\mathbf{O}}
\newcommand{\bP}{\mathbf{P}}
\newcommand{\bQ}{\mathbf{Q}}
\newcommand{\bR}{\mathbf{R}}

\newcommand{\bV}{\mathbf{V}}
\newcommand{\bw}{\mathbf{w}}
\newcommand{\bx}{\mathbf{x}}

\newcommand{\bz}{\mathbf{z}}

\newcommand{\nE}{\mathbb{E}}

\newcommand{\nI}{\mathbb{I}}

\newcommand{\nR}{\mathbb{R}}

\newcommand{\cD}{\mathcal{D}}

\newcommand{\cL}{\mathcal{L}}

\newcommand{\cW}{\mathcal{W}}
\newcommand{\cX}{\mathcal{X}}

\DeclareMathOperator*{\argmin}{argmin~}

\makeatletter
\DeclareRobustCommand\onedot{\futurelet\@let@token\@onedot}
\def\@onedot{\ifx\@let@token.\else.\null\fi\xspace}

\makeatother

\newcommand{\boldparagraph}[1]{\vspace{0.2cm}\noindent{\bf #1:}}

\usepackage{paralist}
\usepackage{epstopdf}
\usepackage{graphicx}
\usepackage{amsmath}
\usepackage{amssymb}
\usepackage{booktabs}
\usepackage{paralist}
\usepackage{multirow}
\usepackage{color, colortbl}
\definecolor{Gray}{gray}{0.97}
\definecolor{Gray8}{gray}{0.9}
\definecolor{Gray18}{gray}{0.8}
\usepackage{subfig}
\usepackage{makecell}

\begin{document}

\title{MaskFuser: Masked Fusion of Joint Multi-Modal Tokenization for End-to-End Autonomous Driving}

\author{Yiqun Duan$^\dag$, Xianda Guo$^\circ$, Zheng Zhu$^\diamond$, Zhen Wang$^\dotplus$, Yu-Kai Wang$^\dag$, Chin-Teng Lin$^\dag$$^\ast$
\thanks{Authors with $^\dag$ are with Human-Centric Artificial Intelligence (HAI) Centre, AAII, University of Techonology Sydney, $^\circ$ with Waytous, $^\diamond$ with Phigent Robotics, $^\dotplus$ with University of Sydney. Author with $^\ast$ are the corresponding author.}
}



\maketitle

\begin{abstract}
Current multi-modality driving frameworks normally fuse representation by utilizing attention between single-modality branches. 
However, the existing networks still suppress the driving performance as the Image and LiDAR branches are independent and lack a unified observation representation. 
Thus, this paper proposes MaskFuser, which tokenizes various modalities into a unified semantic feature space and provides a joint representation for further behavior cloning in driving contexts.
Given the unified token representation, MaskFuser is the first work to introduce cross-modality masked auto-encoder training. 
The masked training enhances the fusion representation by reconstruction on masked tokens.
Architecturally, a hybrid-fusion network is proposed to combine advantages from both early and late fusion:
For the early fusion stage, modalities are fused by performing monotonic-to-BEV translation attention between branches; 
Late fusion is performed by tokenizing various modalities into a unified token space with share encoding on it.
MaskFuser respectively reaches driving score $49.05$ and route completion $92.85\%$ on CARLA LongSet6 benchmark evaluation, which improves the best of previous baselines by $1.74$ and $3.21\%$.
The introduced masked fusion increases driving stability under damaged sensory inputs. 
MaskFuser outperforms ($\Delta$) the best of previous baselines on driving score by $6.55~(27.8\%)$, $1.53~(13.8\%)$, $1.57~(30.9\%)$, respectively given sensory masking ratio $25\%$, $50\%$, and $75\%$. 
\end{abstract}


\section{Introduction}
\label{sec:intro}

Recent trends in autonomous driving tasks could be categorized into pipe-line formation~\cite{gog2021pylot,liu2020hercules,jiao2021greedy,liu2021ground} and End-to-End (E2E) formation~\cite{NVIDIA-2016,chitta2021neat,toromanoff2020end}. 
Pipe-line formation~\cite{li2020lidar, song2020pip,wen2020scenario,claussmann2019review} decompose driving into sequential modules tasks~\cite{zhou2018voxelnet,huang2021bevdet}, localization~\cite{qin2021light,woo2018localization}, scene reconstruction~\cite{dyer2001volumetric,bozic2021transformerfusion,guo2023openstereo,duan2023diffusiondepth}, planning~\cite{letchner2006trip,abu2022comprehensive,duan2024prompting}, and control~\cite{tsugawa1994vision,kong2015kinematic}. 
E2E driving~\cite{hu2022st,tampuu2020survey} applies state-to-action imitation~\cite{NVIDIA-2016} or reinforcement learning~\cite{toromanoff2020end} on mediate feature representation states to teach agents to behave properly given driving context. 

Multi-modality sensory fusion is first introduced~\cite{transfuser,zhang2022mmfn} by pipeline methods to enhance the performance of specific perception tasks, such as 3D object detection~\cite{huang2021bevdet,liu2022bevfusion}, depth estimation~\cite{li2022bevdepth,zhao2020monocular}, and instance motion forecasting~\cite{hu2021fiery,luo2018fast}. 
These methods utilize tight geometrical correspondence of modalities, 
and perform fusion by projecting features into a geometrically unified feature space, where, Bird’s Eye View (BEV)~\cite{fadadu2022multi,chen2017multi,zhou2020end,huang2021bevdet,liu2022bevfusion,li2022bevdepth} emphasizes on ego relations and Range View (RV)~\cite{meyer2020laserflow,li2020end,xu2019depth,meyer2019lasernet} emphasizes on the semantic perception.
However, these methods are not perfectly suitable for comprehensive E2E driving tasks. 
ST-P3~\cite{hu2022st} proposes to absorb architectural prior knowledge by multi-stage geometrical perception and motion prediction. 
TransFuser~\cite{transfuser} observes that pure geometrical fusion hampers the performance of comprehensive and complex urban E2E driving, as the geometrical translation may lose key information for driving 
\footnote{eg., Traffic light color in far distance may only affect several pixels in BEV feature after several convolutions.}.
Thus, TransFuser~\cite{transfuser}, followed by, MMFN~\cite{zhang2022mmfn}, and Interfuser~\cite{shao2022safety} present a bi-stream structure, which utilizes independent CNN~\cite{he2016deep,duan2019learning} branches respectively for camera and LiDAR.
The fusion is realized by applying element-wise attention between modalities through transformer layers. 
However, these methods are actually \textit{feature exchange} through attention from different modalities, where the joint-modality representation and the feature alignment have not been investigated. 


To further enhance the joint feature representation, we propose \textbf{MaskFuser}, a hybrid multi-modality fusion framework that obtains joint representation by tokenizing various modalities into a unified semantic space. 
The \textit{hybrid} formation denotes MaskFuser combines the \textit{early fusion} and \textit{late fusion} sequentially as shown in Fig.~\ref{fig:HybridNet}. 
First, separated encoder branches are applied to the image and LiDAR modality to extract lower-level features.
Instead of presenting element-wise attention as previous works~\cite{transfuser,zhang2022mmfn}, MaskFuser introduces a Monotonic-to-BEV Translated (MBT) attention to perform the \textit{early fusion}.
MBT attention is applied between separated early branches to enrich the feature quality by geometrical projection.
Then, the fused features are tokenized into tokens and concatenated into a unified token space. 
We propose to perform \textit{late fusion} on these tokens from different modalities by applying a share transformer~\cite{wang2022continual,duan2022position,dosovitskiy2020image} encoder. 
By modeling modality tokens as language words, the proposed tokenized share encoding enhances joint representation rather than merely applying feature exchanges between branches~\cite{transfuser,nie2020mmfn}. 

Given joint token representation, MaskFuser further introduces masked reconstruction pretraining~\cite{he2022masked} (Fig.~\ref{fig:HybridNet}).
We randomly masked $75\%$ tokens before the shared encoder, where tokens from each modality have an even chance to be masked. 
The visible tokens are forced to reconstruct the complete joint token sequence by considering both spatial relations and cross-modality connections inside the token space. 
The joint token sequence is supervised by predicting the original multiple sensory inputs. 
Meanwhile, the masked tokens are supervised by auxiliary tasks to predict BEV map~\cite{liu2022bevfusion}, semantic segmentation map~\cite{li2022uncertainty}, and depth map~\cite{li2022bevdepth} given partially visible tokens. 


MaskFuser enhances the representation for driving context in threefold: 
1) Joint token representation with share encoding align multiple modalities into a unified semantic space and bring deeper feature integration. 
2) Masked token reconstruction forces encoders to keep rich details in joint representation, which is crucial for driving imitation in complex urban environments. 
3) Training on partially visible tokens increases driving stability under sensory-damaged conditions. 
On LongSet6 benchmark evaluation in the CARLA simulator, MaskFuser reaches driving score $49.05$ and route completion $92.85\%$, which outperforms previous fusion methods by $2.10$ and $3.21\%$.
Given partially damaged sensory inputs, MaskFuser outperforms previous baselines by
($\Delta$) $6.55~(27.8\%)$, $1.53~(13.8\%)$, $1.57~(30.9\%)$, given sensory masking ratio $25\%$, $50\%$, and $75\%$ respectively.
The ablation study provides a detailed analysis of each component of MaskFuser as well as a discussion of early and late fusion. 
The contribution of MaskFuser could be categorized into threefold:
\begin{compactitem}
\item MaskFuser is the first work to propose masked fusion in a driving context. It enhances the perception details and the driving stability under damaged sensory inputs by masked token reconstruction training.
\item MaskFuser proposes a \textit{hybrid} network with monotonic to BEV translation attention and share transformer encoder on unified token representation for E2E driving.
\item Experimental results suggest MaskFuser could improve the driving sore (DS) and route completion (RC) by $2.10$ $(4.5\%)$ and $3.21\%$ $(3.6\%)$. During masked $75\%$ sensory testing, the improvements on DS and RC are $1.57~(30.9\%)$ and $3.15\%$ $(29.2\%)$ respectively.
\end{compactitem}

\section{Related Works}
\textbf{Imitation Learning for End-to-End Driving:}
As for End-to-End autonomous driving, previous works presents a \textit{destructured, then structured} formation. 
Early explorations, such as ALVINN~\cite{pomerleau1988alvinn}, DAVE~\cite{bojarski2016end} models simple projection relationship between input single view camera and steer wheel or accelerator angle. 
At this stage, the popular formation is to remove human prior knowledge and give the network a more direct learning target to increase the fitting ability for fully supervised behaviour cloning. 
Still, the sensory input is limited at early era, while multiple imitation learning techniques~\cite{codevilla2018end, bansal2018chauffeurnet,kendall2019learning} are introduced to improve the data quality for behavior cloning. 
Subsequently, utilization of camera sensors are extended \cite{hecker2018end} from single view to multiple views to increase the planning stability. 
Latest works start to re-introduce prior knowledge to improve the driving performance.  
WOR \cite{wor} and MARL\cite{marl} suggest that introducing prior trained knowledge from related vision tasks such as detection~\cite{resnet,CNN,yolov3} could improve the model performance. 
NEAT~\cite{chitta2021neat} introduced attention module between different camera views to enhance the feature quality.
ST-P3~\cite{hu2022st} combine human prior module design from pipe-line methods to increase the driving rationality. 

\textbf{Multi-Modality Fusion For Driving Tasks:}
However, monotonic camera sensor is easily affected by lighting condition and occlusion. Later works~\cite{pointpillars,rhinehart2019precog,vectornet,lbc} extend the sensory modalities to LiDAR, Radar and Openmap to increase the driving stability. 
Multi-Modality fusion naturally arouse attention.
Most previous approaches of modality fusion are designed for specific perception tasks like such as 3D object detection~\cite{huang2021bevdet,liu2022bevfusion}, depth estimation~\cite{li2022bevdepth,zhao2020monocular}, and instance motion forecasting~\cite{hu2021fiery,luo2018fast}. 
Especially for object detection, the introduction of human selected mediate representation, BEV feature map, has largely increase the average precision on driving scenario. 
BEVFusion~\cite{liu2022bevfusion} fuse LiDAR with Image features by using completely independent encoders with geometric projection into same BEV feature space at late stage. 
However, Transfuser~\cite{transfuser} observed that pure geometrical fusion representation hampers the performance of comprehensive and complex urban E2E autonomous driving. 
MMFN~\cite{zhang2022mmfn} follows the fusion strategy and extend modality to Radar and OpenMap with early fusion strategy. 
UniAD~\cite{hu2023planning} proposed to leverage multiple pipeline modules into a unified autonomous driving network, combining advantages of the both the end-to-end and pipeline methods. 
GenAD~\cite{zheng2024genad} further introduced tokenized latent space learning and generative prediction based on UniAD. 
None of previous methods have explored using masked image model to enhance the feature quality of the joint representation. The proposed method could provide better feature extractors for these driving frameworks. 

\section{Methodology}

\begin{figure*}[!t]
  \centering
  \includegraphics[width=0.92\textwidth]{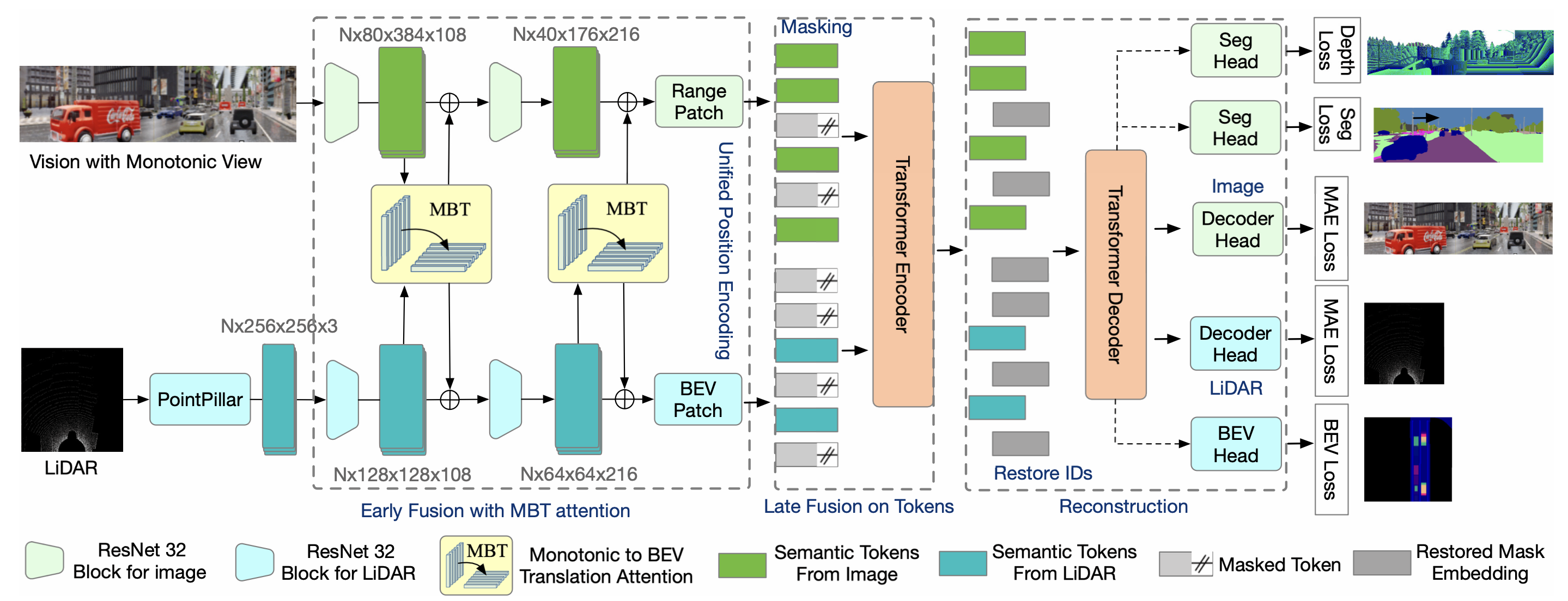}
  \caption{Overall network structure of MasFuser with hybrid fusion structure for pretraining. The network applies Monotonic-to-BEV Translation (MBT) Attention for early fusion. Features from various modalities are patched into tokens with unified position encoding and masked randomly. A shared transformer encoder is applied to get the perception state of the current environment. 
  For masked pretraining, an additional transformer decoder is applied to reconstruct both original Camera $\&$ LiDAR sensory inputs and auxiliary tasks. 
  }
  \label{fig:HybridNet}
\end{figure*}

\subsection{Overview}\label{subsec:problemset}

\textbf{Problem Setup:} 
MaskFuser follows the previous widely accepted setting of E2E driving~(\cite{toromanoff2020end,transfuser,zhang2022mmfn}) that the goal is to complete a given route while safely reacting to other dynamic agents, traffic rules, and environmental conditions. 
Thus, the goal is to learn a policy $\mathbf{\pi}$ given observation.
As MaskFuser emphasizes more on multi-modality fusion, we choose the \textbf{Imitation Learning (IL)} approach to learn the policy. The goal is to obtain policy  $\mathbf{\pi}$ by imitating the behavior of an expert $\pi^{*}$. 
Given an expert, the learning dataset $\cD = \{ (\cX^i, \cW^i) \}$ could be collected by letting the expert perform similar routes, where $\cX^i = \{(\bx_{im}^i, \bx_{Li}^i)_t \}_{t=1}^T$ denotes image and LiDAR sensory observations of the current state, and $\cW = \{ (x_t, y_t) \}_{t=1}^T$ denotes the expert trajectory of waypoints. 
Here, $x_t, y_t$ denotes the 2D coordinates in ego-vehicle (BEV) space.
Thus, the learning target could be defined as in Eq.~\ref{eq:target}.
\begin{equation}\label{eq:target}
 \argmin_{\pi} \nE_{(\cX, \cW) \sim \cD} \left[ \cL_{wp} (\cW, \pi(\cX)) \right]
\end{equation}
where $\cL_{wp}$ is the waypoint loss defined in Eq.~\ref{eq:waypointsloss_sup}, and $\pi(\cX)$ is the predicted waypoints given observation $\cX$ through policy $\pi$ to be learned. 

\textbf{Formation:}
In this paper, the policy $\pi(\cX)$ is realized by the combination of a hybrid fusion network (Sec.~\ref{subsec:hybridfusion}) and a waypoints prediction network (Sec.~\ref{subsec:waypoint}), where the fusion network transfer multi-modality sensory inputs $\cX$ into semantic tokens $\bF_s$, and waypoints network predicts future goal points $\cW$ given $\bF_s$. 
Cross-modality masking (Sec.~\ref{subsec:cmmae}) is applied on semantic tokens $\bF_s$ to further enhance the feature quality by masked sensory reconstruction. 
Then a PID Controller $\nI$ (Sec.~\ref{subsec:controller}) is applied on the waypoints $\cW$ decision and decomposed it into practical control, ie., steer, throttle, and brake, through $\ba = \nI (\cW)$. 

\subsection{Hybrid Fusion Network}\label{subsec:hybridfusion}
This section gives an architectural overview of MaskFuser. 
MaskFuser proposed a hybrid network shown in Fig.~\ref{fig:HybridNet} that combines the advantages of \textit{early fusion} and \textit{late fusion}. 
The network consists of two stages. 

\textbf{Early Fusion:} At the first stage, we apply two separate CNN branches to extract shallow features respectively from monotonic image and LiDAR inputs. 
For the image branch, MaskFuser concatenates three front view camera inputs each with 60 Fov into a monotonic view and reshaped into shape $3\times160\times704$.
For the LiDAR branch, MaskFuser reprocesses the raw LiDAR input with PointPillar~\cite{pointpillars} into BEV feature with shape $33\times256\times256$.
Since the lower-level features still retain strong geometric relations, the separated encoder could extract tight local feature representation with fewer distractions. 
%
A novel monotonic-to-BEV translation (MBT) attention is applied to enrich each modality with cross-modality assistance. 
MBT attention translates both images and LiDAR features into BEV feature space and performs a more precise spatial feature alignment compared to previous element-wise approaches. 

\textbf{Late Fusion:} At the second stage, the network respectively tokenizes~\cite{dosovitskiy2020image} feature maps from Image and LiDAR stream into semantic tokens, respectively denoted by green and blue in Fig.~\ref{fig:HybridNet}. 
The late fusion is performed by directly applying a shared transformer encoder over the concatenated token representation. 
The shared encoder with position embedding could force the tokens from various modalities aligned into a unified semantic space. 
Also, by treating multi-sensory observation as semantic tokens, we could further introduce masked auto-encoder training mentioned below. 


\subsubsection{Early Fusion: Monotonic-to-BEV Translation (MBT)}\label{subsec:MBT}
attention performs cross-modality attention more precisely by introducing human prior knowledge (BEV transformation). 
Inspired by Monotonic-Translation~\cite{saha2022translating}, we model the translation as a sequence-to-sequence process with a camera intrinsic matrix. 
\begin{figure}[t]
  \centering
  \vspace{-10pt}
  \includegraphics[width=0.35\textwidth]{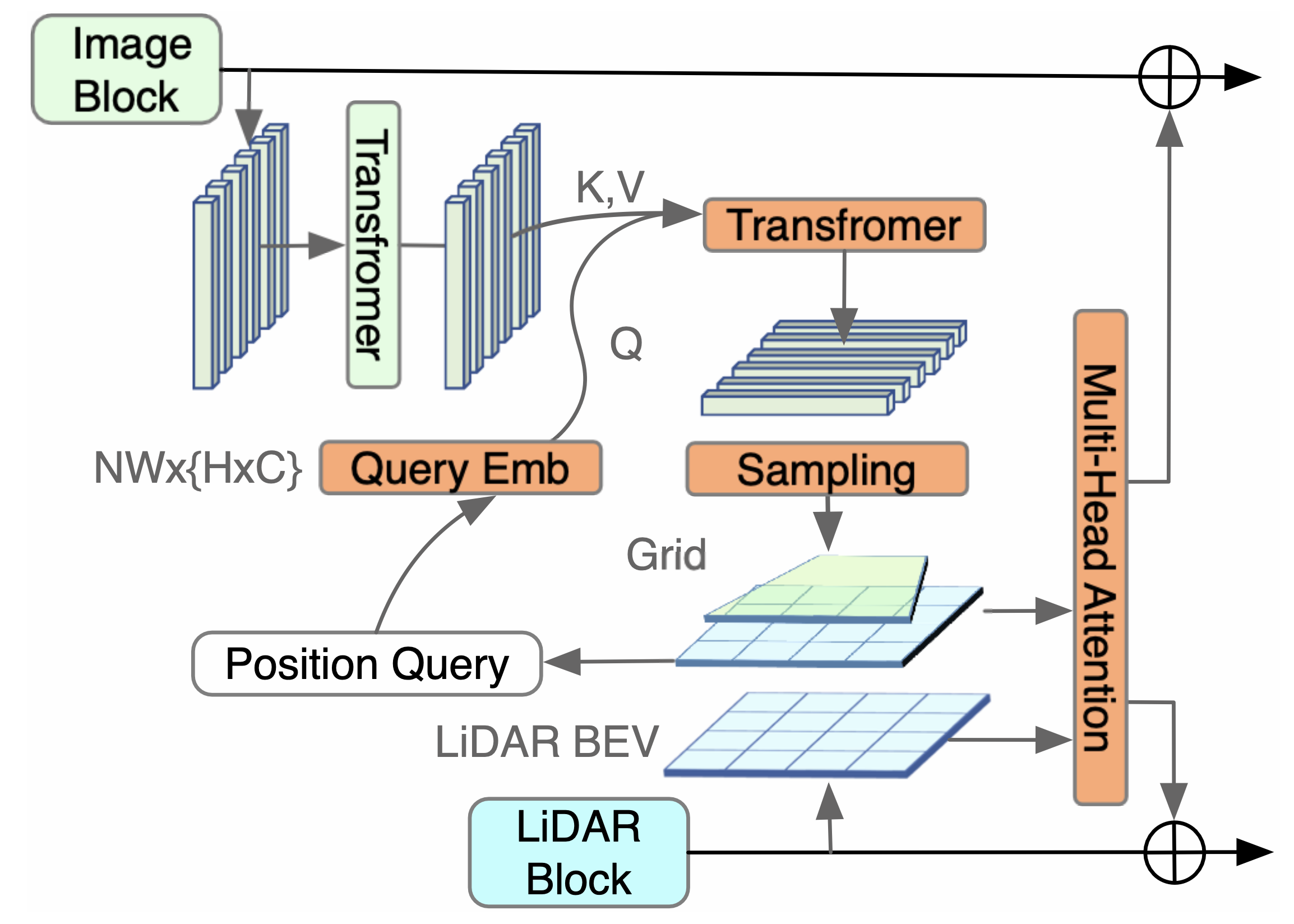}
  \caption{The structure of the MBT attention module, where the features from monotonic view are projected into BEV space through a sequence-to-sequence formation.  
  }
  \label{fig:mbt}
  \vspace{-10pt}
\end{figure}
The detailed structure of MBT attentions is shown in Fig.~\ref{fig:mbt}, where the feature map from image stream with shape $\bF_{im} \in \mathbb{R}^{N\times C\times H \times W}$ is reshaped along width dimension $W$ into image columns $\bF_c \in \mathbb{R}^{NW \times \{H \times C\}}$. 
The column vectors are projected into a set of mediate encoding $\{\bh_i \in \mathbb{R}^{ H \times C}\}_{i=1}^{NW}$ through a transformer layer with multi-head self-attention.  
We treat mediate encoding of monotonic view $\bh_i$ which projects the key and value representing the range view information to be translated. 
\begin{equation}
    K(\bh_i) = \bh_i W_K, \quad V(\bh_i) = \bh_i W_V
\end{equation}
A grid matrix is generated indicating the desired shape of the target BEV space.   
We generate position encoding $\{g_i \in \mathbb{R}^{r \times C }\}_{i=1}^{NW}$ along side each radius direction inside the grid with depth $r$. 
The grid position $g_i$ is tokenized into query embedding and query the $\bh_i$ with radius directions~\footnote{The radius coordinates are calculated by given camera intrinsic matrix, FOV, and prefixed depth length.} as defined in Eq.~\ref{eq:translate} and generate the translated feature map $\bF_{mbt}$.
\begin{equation}\label{eq:translate}
    \begin{aligned}
    \rm{s}(\bg_i, \bh_j) &= \frac{ Q(\bg_i) K(\bh_j)^T }{\sqrt{D}}, \quad Q(\bg_i) = \bg_i W_Q \\
    \bF_{mid} &= \sum _{i} \frac{\rm{exp}(\rm{s}(\bg_i, \bh_j))}{\sum _{j} \rm{exp}(\rm{s}(\bg_i, \bh_j))} V(\bh_i)
    \end{aligned}
\end{equation}
where $\rm{s}(\bg_i, \bh_j)$ is the scaled dot product~\cite{vaswani2017attention} regularized by dimension $D$. 
Yet, since the monotonic view only suggests information inside certain FOV (Field of view), we apply a sampling process $\bF_{mbt} = \rm{P}(\bF_{mid})$ to sample points inside FOV decided by camera intrinsic matrix into BEV feature map $\bF_{mbt}$. 
The translated feature map $\bF_{mbt}$ and feature map from LiDar $\bF_{Li}$ is reshaped by flatten along width $W$ and depth $r$ into vectors and concatenate into sequence $\bF_{in} = \rm{cat}(\bF_{mbt}, \bF_{Li})$. 
The transformer layer is applied on $\bF_{in} \in \mathbb{R}^{N^\star \times C}$ to perform self multi-head attention~(\cite{vaswani2017attention}) between each token in $N^\star$ dimension.

\subsubsection{Late Fusion: Unified Tokenization}\label{subsec:mmtransfuser}

After the early fusion network, we intend to align features from both modalities into a unified semantic token space, where we treat each $16\times 16 \times C$ feature segment to be a semantic \textit{word} as shown in Fig.~\ref{fig:HybridNet}. 
To maintain the spatial relationships across these tokens, we use the 2d position encoding~\cite{dosovitskiy2020image}.
A segment embedding is added with positional embedding $\mathbf{PE}+\mathbf{SE}$ to indicate the position information is from range view or BEV feature space.
The feature map from different modalities is tokenized and concatenated into unified tokens $\bF_{s} \in \mathbb{R}^{N^\star \times C}$ with patch size $16\times16$ by adding $\mathbf{PE}+\mathbf{SE}$ to the token. 
A 4-layer transformer encoder is applied on the unified token sequence $\bF_{s}$ by setting query $Q(\bF_{s})$, key $K(\bF_{s})$, and value $V(\bF_{s})$. 
As the late fusion is performed by share transformer encoding, MaskFuser could introduce a deeper feature interaction between each semantic token layer by layer. 
The late fusion outputs the fused feature map $\bF_{f}$ sequence for further waypoints prediction.

\subsection{Masked Cross-Modality Pretraining }\label{subsec:cmmae}

The late fusion is enhanced by masked auto-encoder (MAE~\cite{he2022masked}) reconstruction pretraining as shown in Fig~\ref{fig:HybridNet}.
The \textit{masked} fusion strategy benefits the driving task threefold:
1) The masked cross-modality reconstruction on joint tokens introduces \textit{self-supervised learning for perception}. It could largely increase the training data with raw sensory inputs. 
2) The reconstruction forces the joint representation to keep rich details, which is crucial for driving imitation in complex urban environments. The details such as small traffic lights in the distance could help overcome failure cases of ignoring red light~\cite{transfuser}. 
3) Training on partially visible tokens increases driving stability under sensory-damaged situations. 


The masking is applied on multi-modality joint feature map $\bF_s \in \mathbb{R}^{N^\star \times C}$ with mask ratio $r$ and records the mask position with restore ids. 
Since we use direct transformer structure~\cite{dosovitskiy2020image}, we only send visible feature tokens to the encoder. 
After the encoder processes the visible feature into $\bF_f^{vis} \in \mathbb{R}^{{(1-r)N^\star} \times C}$, the feature is completed back to full length by filling mask embedding at blank positions given restore ids as shown in Fig.~\ref{fig:HybridNet}. 
Based on the feature map after completion, a decoder transformer~\cite{he2022masked} is applied to produce decoder feature tokens $\bF_{dec}$. 
Given decoder feature $\bF_{dec}$, we split the decoder feature by position as $\bF_{dec}^{image}, \bF_{dec}^{Li} = \rm{pos}(\bF_{dec})$ into restored image tokens and LiDAR tokens. 
Two MLP layers are respectively applied on two sets of tokens to reconstruct the original sensory input as defined in Eq.~\ref{eq:mse}.
\begin{equation}\label{eq:mse}
    \begin{aligned}
    \Bar{\bx}_{im} = \rm{mlp}(\bF_{dec}^{image}), \ \Bar{\bx}_{Li} = \rm{mlp}(\bF_{dec}^{Li})  \\
    \cL_{\rm{mse}} = \lambda_1 \sum_i^N(\bx_{im} - \Bar{\bx}_{im})^2 + \lambda_2 \sum_i^N(\bx_{Li} - \Bar{\bx}_{Li})^2
    \end{aligned}
\end{equation}
where $\Bar{\bx}_{im}$ and $\Bar{\bx}_{Li}$ denotes the predicted sensory input through decoder. 
The masked auto-encoder training is unsupervised, thus MaskFuser could utilize a larger training set to enhance the feature quality.

\subsection{Auxiliary Tasks:} 
\label{ap:auxiliarytask}

\begin{figure}[hbpt]
  \centering
  \includegraphics[width=0.45\textwidth]{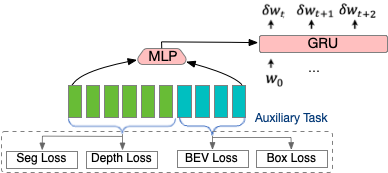}
  \caption{The structure of waypoints prediction network, where dotted line denotes the auxiliary loss. 
  }
  \label{fig:waypoints}
\end{figure}

We follow the basic setting for End-to-End driving tasks that introduce auxiliary tasks to increase understanding of complicated driving scenarios. We introduce technology details of auxiliary tasks incorporating the training target in the main paper.  
We split the semantic tokens according to the original position into image tokens and BEV tokens as shown in Fig.~\ref{fig:waypoints}. 
The auxiliary tasks are respectively applied to the two kinds of tokens. 

\boldparagraph{Image Tokens} 
The image tokens contain rich information from the range view.
The image tokens are supervised by combining 2D depth estimation and 2D semantic segmentation as auxiliary tasks have been an effective approach for image-based end-to-end driving.
We use the same decoder architecture as the Transfuser baseline of \cite{transfuser, neat} to decode depth and semantics from the image branch features. The depth output is supervised with an $L_1$ loss, and the semantics with a cross-entropy loss. For the segmentation task, it contains 7 semantic classes: (1) unlabeled, (2) vehicle, (3) road, (4) red light, (5) pedestrian, (6) lane marking, and (7) sidewalk.

\boldparagraph{BEV Tokens} 
We predict both the BEV segmentation and BEV bounding boxes prediction task on the BEV tokens. 
The BEV tokens are restored to a complete feature map according to the original position provided before tokenizing. 
For BEV segmentation, we predict a three-channel prediction task containing the classes road, lane marking, and others. 
This encourages the intermediate features to encode information regarding drivable areas. The map uses the same coordinate frame as the LiDAR input. 
The prediction output is obtained from the feature map of the convolutional decoder on the restored feature map. 
The prediction map is with the size of $64 \times 64$ for computational efficiency. 

The BEV tokens are further enhanced by vehicle detection tasks. 
CenterNet decoder~\cite{duan2019centernet} is used as the detection head. 
Specifically, it predict a position map $\hat{\bP} \in [0,1]^{64 \times 64}$ from the restored BEV feature map. 
The 2D target label for this task is rendered with a Gaussian kernel at each object center of our training dataset. 
Following existing works~\cite{transfuser, mousavian20173d}, the detection is decomposed into two stages with coarse to fine formation.
To predict the coarse orientation, we discretize the relative yaw of each ground truth vehicle into 12 bins of size 30$^{\circ}$ and predict this class via a 12-channel classification label at each pixel, $\hat{\bO} \in [0,1]^{64 \times 64 \times 12}$, as in~\cite{yang20203dssd}. Finally, we predict a regression map $\hat{\bR} \in \nR^{64 \times 64 \times 5}$. This regression map holds three regression targets: vehicle size ($\in \nR^2$), position offset ($\in \nR^2$), and orientation offset ($\in \nR$). The position offset is used to make up for quantization errors introduced by predicting position maps at a lower resolution than the inputs. The orientation offset corrects the orientation discretization error~\cite{yang20203dssd}. 
Note that only locations with vehicle centers are supervised for predicting $\hat{\bO}$ and $\hat{\bR}$. The position map, orientation map, and regression map use a focal loss, cross-entropy loss, and $L_1$ loss respectively.

\subsection{Waypoints Prediction}\label{subsec:waypoint}

The waypoints prediction is applied to joint token representation.
Different from previous works~\cite{transfuser,zhang2022mmfn} which aggregate features by adding features from modalities, we apply an MLP layer to project joint tokens into observation state $\bz \in \mathbb{R}^{256}$. 

An auto-regressive GRU~\cite{chung2014empirical} module is applied as the prediction network, where the observation state $\bz$ is utilized as the hidden state. 
GPS coordinates of the target location (transferred into current ego coordinates) are concatenated with the current location as the network inputs. 
At each frame, the GRU network predicts the differential waypoints $\{ \Delta \bw_t \}_{t=1}^T$ for the next T time steps and obtain future waypoints by $\{\bw_t = \bw_{t-1} + \Delta \bw_t \}_{t=1}^T$. 
The waypoints loss function is defined in Equation~\ref{eq:waypointsloss_sup}
\begin{equation}\label{eq:waypointsloss_sup}
    \cL_{wp} = \sum_{t=1}^T \left\|\bw_t - \bw_t^{gt}\right\|_1
\end{equation}
where the $\bw_t^{gt}$ is the ground truth from the expert route. 
MaskFuser adds auxiliary tasks (appendix~\ref{ap:auxiliarytask}) to increase the interoperability of the joint tokens. 
The training targets is constructed $\cL_{total}$ in Eq.~\ref{eq:combined_loss_sup} by applied weighted summation on waypoint prediction $\cL_{wp}$ and auxiliary losses $\{\cL_{i}\}$. 
\begin{equation}\label{eq:combined_loss_sup}
    \cL_{total} = \lambda_{wp}\cL_{wp} + \sum_i \lambda_{i}\cL_{i}, i \in {tasks}
\end{equation}
The joint tokens are separated back into image tokens and LiDAR tokens in implementation.
For image tokens, Depth prediction~\cite{laina2016deeper} and Semantic Segmentation~\cite{treml2016speeding} tasks are applied. 
For BEV tokens, object detection~\cite{zhang2019freeanchor} and HD Map prediction~\cite{pointpillars} are applied.

\subsection{Controller}\label{subsec:controller}

As the emphasis of MaskFuser is the feature fusion part, we follow the basic setting of most previous end-to-end works~\cite{transfuser,zhang2022mmfn}, where the predicted waypoints are converted into steer, throttle, and brake values through a set of PID controllers.
The turn controller takes the weighted average of the vectors
between waypoints along the consecutive time steps and calculated the steer angles. 
Also, a speed controller takes the distance between the sequence of history and previous waypoints and calculated the throttle and brake values based on the PID smoothed speed value. 
Similar to previous methods, in case of the car is trapped under certain circumstance, MaskFuser applies creeping strategy. 
If the car is detected to stay at the same location for a long time, the controller will force the car to move and change the observation as well as the care status at a very low speed (4m/s). 
As MaskFuser is to enhance observation features to improve the end-to-end driving task, we only apply the basic safety heuristic rules~\cite{transfuser} to make it a fair comparison.
It is noted that the driving performance could be further improved by adding more safety rules.

\section{Experiments}

The experimental setting and the implementation details are respectively introduced in Section~\ref{exp:setting} and Section~\ref{exp:implementation}
The driving performance discussed and Section~\ref{exp:drivingscore}. We extend the driving evaluation to the damaged perception condition in Section~\ref{exp:mae}.
The ablation study is conducted in Section~\ref{exp:ablation} to discuss each component of MaskFuser. 

\subsection{Experimental Setting}\label{exp:setting}

\noindent\textbf{Simulation Environment:} MaskFuser perform the end-to-end driving task under the simulator. 
We use the \textbf{CARLA Simulator}~\footnote{\url{https://carla.org/}}, a high-performance driving simulator under urban scenarios as our simulation environment. 
The driving is performed in a variety of areas under various weather and lighting condition. 
The agent is expected to perform driving given predefined routes under complicated traffic and weather conditions, where the routes are a sequence of sparse goal locations with GPS coordinates. 

\noindent\textbf{Data Collection:}
As mentioned in Sec.~\ref{sec:intro}, the driving agent is trained through imitation learning, and the training data is collected by letting the expert agent drive inside the simulator. 
We use CARLA version 0.9.10 to collect 1000 routes in 8 officially provided town maps with an average length of 400m by using CARLA rule-based expert auto-agent~\cite{transfuser} with 228k training pairs. 
As MaskFuser applies an unsupervised masked auto-encoder pre-train without requiring expert route actions, we collect additional 200k unsupervised training data, with only perception results under various weather $\&$ lighting conditions. 

\noindent\textbf{Evaluation Benchmark:}
The MaskFuser is evaluated by both online benchmark and offline benchmarks. 
Although the CARLA simulator provides an official evaluation leaderboard, the use time is restricted to only 200 hours per month, which makes the official leaderboard unsuitable for ablation studies or obtaining detailed statistics involving
multiple evaluations of each model. 
We conduct our detailed ablation and comparison based on the Longeset6 Benchmark proposed by Transfuser~\cite{transfuser}, which chooses the 6 longest routes
per town from the officially released routes from the carla challenge 2019 and shares quite a similarity with the official evaluation. 

\noindent\textbf{Evaluation Metrics:}
For both online evaluation and offline evaluation, we follow the official evaluation metrics to calculate three main metrics, \textbf{Route Completion (RC)}, \textbf{Infraction Score (IS)}, and \textbf{Driving Score (DS)}.
The RC score is the percentage of route distance
completed. Given $R_i$ as the completion by the agent in route $i$, RC is calculated by averaging the completion rate $RC = \frac{1}{N} \sum_i^N R_i$. 
The IS is calculated by accumulating the infractions $P_i$ incurred by the agent during completing the routes. 
The driving score is calculated by accumulating route completion $R_i$ with infraction multiplier $P_i$ as $DS=\frac{1}{N} \sum_i^N R_i P_i$. 
We also calculate the detailed infraction statistical details according to the official codes. 

\subsection{Implementation Details}\label{exp:implementation}
\boldparagraph{Hybrid Fusion Network}
MaskFuser also uses the two sensory inputs as modalities, cameras, and  LiDAR as the sensory inputs. 
For camera input, we concatenate three monotonic camera inputs into a narrow range view with FOV 120 degrees and reshape the sensory input into shape $(160, 704)$. 
The LiDAR point cloud is converted to BEV representation~\cite{pointpillars} and reshaped into $256, 256$. 
For CNN down-sampling blocks, we use ImageNet pre-trained RegNet-32~\cite{xu2022regnet} for both Image and BEV LiDAR branches. 
Similar to Transfuser~\cite{transfuser}, we apply angular viewpoint augmentation for the training data, by randomly rotating the LiDAR inputs by ±20◦ and adjusting ground truth labels accordingly.

MBT modules are applied between feature maps after the first two convolutional blocks, where the resolution is respectively $(C_1, 40, 176)$ and $(C_2, 20, 88)$, where $C_1, C_2$ denotes the channel dimension $72,216$ under our setting. 
For the MBT module, two transformer layers with hidden dimensions $512$ and $4$ attention heads are respectively applied as encoder and decoder sequentially.
For polar ray grid sampling details, we follow the settings of previous work~\cite{saha2022translating}.   
The two MBT module make attention to different depths with position query shape $(C, 25, 176)$ and $(C, 13,176)$, where the on-depth range $0\%-62.5\%$ $62.5\%-95\%$ of predefined max depth $z$. 
Given the setting on BEV view, we define $8$ pixel represents $1.0$ meters in real-life scenarios, thus MBT attention covers the monotonic depth range accordingly up to $30.5$ meters. 

The feature maps are respectively with resolution $20\times88\times512$ and $32\times32\times512$ after two times sampling and MBT fusion.
MaskFuser slice both feature maps into $4\times4$ patches and concatenate both features into the unified sequence of tokens. 
Thus, the MaskFuser respectively has $5\times22$ image tokens and $8\times8$ BEV tokens. 
For the late fusion part, a unified Vit encoder is applied on the semantic token sequence with length $174$. 
The Vit encoder has four layers of transformer layers, each with 4 attention heads.
We directly use the channel number ($512$) after the early fusion as the dimension of the tokens for transformer encoders. 

\boldparagraph{MAE Training for MaskFuser}
The feature maps are respectively with resolution $20,88$ and $32,32$ after two times sampling and MBT fusion.
MaskFuser slice both feature maps into $4\times4$ patches and concatenate both features into the unified sequence of tokens. 
Thus, the MaskFuser respectively have 

For pretraining, we randomly mask $75\%$ of the tokens and record the masked ids, and only send $25\%$ visible tokens into encoders. The encoded semantic tokens are completed by filling missing tokens with a mask embedding given recorded masked ids back to their original shape and are sent into the decoder to reconstruct the original sensory inputs. 
The encoder and decoder structure are vision transformers~\cite{dosovitskiy2020image} each with 4 standard transformer layers.
During the pretraining period, we apply the same auxiliary tasks as mentioned in appendix~\ref{ap:auxiliarytask} to force the model to predict perception results based on partially visible semantic tokens.

\boldparagraph{Driving Imitation Training}
We train the models with 4 Quadro RTX8000 GPUs for 41 epochs. 
We use an initial learning rate of
$1e-4$ and a batch size of 32 per GPU, and disable batch
normalization in the backbones during training. 
We reduce the learning rate by a factor of 10 after epoch 30. 
For all models, we use the AdamW optimizer, which is a variant of Adam. 
Weight decay is set to 0.01, and Adam beta values to the
PyTorch defaults of 0.9 and 0.999.
The combination weights of different losses are respectively set to $1.0$ for BEV segmentation, semantic segmentation, BEV prediction, and waypoints prediction. 
Other assistant losses such as detection loss, velocity prediction loss, and yaw angle classification loss are with a weight of $0.2$.
The final loss is obtained from the weighted combination of these losses. 
We directly load the weights from pretraining to initialize the network weights.

\subsection{Driving Performance}\label{exp:drivingscore}

\textbf{Comparison Baseline:} We compare MaskFuser with state-of-the-art (SOTA) E2E driving baselines below: 1) NEAT~(\cite{chitta2021neat}) denotes attention field on image inputs.
2) WOR~(\cite{chen2021learning}), multi-stage Q-function based learning framework; 3) GRIAD~(\cite{chekroun2021gri}) general reinforced imitation baseline; 4) LAV~(\cite{chen2022learning}), learning from multiple observed agents; 5) Transfuser~(\cite{transfuser}) and InterFuser~(\cite{shao2022safety}), respectively pure fusion baseline and safety assisted baseline. 
We also include Late-Transfuser (Late TF) and geometrical fusion (GF) from~\cite{transfuser} to report the early-late fusion comparison. 

\begin{table}[hbpt]
\setlength\tabcolsep{5pt}
\label{tb:longset6}
\caption{ Driving Evaluation on LongSet6 on Carla 0.9.10}
\centering
\resizebox{1\columnwidth}{!}{
\begin{tabular}{clccc}
\toprule
\textbf{Safety} &                 & \textbf{DS $\uparrow$}          & \textbf{RC    $\uparrow$ $\scriptsize{(\%)}$}      & \textbf{IS$\uparrow$}       \\ \midrule
&NEAT            &  20.63±{2.34}    & 45.31±{3.25}  &   0.54±{0.03}          \\
&WOR              & 21.48±{2.09}  &  52.28±{4.07} &   0.51±{0.05}  \\
&GRIAD           & 26.98±{4.23}  & 71.43±{3.57}      &    \underline{0.56±{0.04}}   \\
&LAV     & 37.62±{1.77}    &  83.94±{2.69}  &   \textbf{0.59±{0.05}}        \\ 
&GF  &   27.44±{2.11}  &   80.43±{3.98}   &    0.32±{0.02}         \\
&Late TF  &   35.42±{4.07}  &   83.13±{1.01}   &    0.39±{0.04}         \\
&TransFuser       & \underline{46.95±{5.81}} & \underline{89.64±{2.08}} & 0.52±{0.08} \\
&MaskFuser        &  \textbf{49.05±{6.02}}   &  \textbf{92.85±{0.82}}  &   \underline{0.56±{0.07}}        \\ \midrule

\rowcolor{Gray8} \checkmark & InterFuser       & \underline{49.86±{4.37}} & \underline{91.05±{1.92}} & \underline{0.60±{0.08}} \\
\rowcolor{Gray8} \checkmark & MaskFuser \small{+ InterFuser}  &  \textbf{50.63±{5.98}}   &  \textbf{92.89±{0.80}}  &   \textbf{0.62±{0.08}}      \\ \midrule
\rowcolor{Gray18} & Expert           &  75.83±{2.45}   &   89.82±{0.59}  &  0.85±{0.03}      \\ \bottomrule
\end{tabular}
}
\end{table}

The performance is reported in the offline benchmark LongSet6~\cite{transfuser} in Table~\ref{tb:longset6}. 
The mean and variance metric values are calculated by evaluating each agent three times. 
MaskFuser reach mean driving score (DS) $49.05$ and route completion (RC) $92.85$, which improves previous best results without safety controller by $2.10(+4.5\%)$ and $3.21\%(+3.6\%)$.
Meanwhile, the variance of RC reduced from $2.08\%$ to $0.82\%(-60.5\%)$ which suggests the stability improvement of MaskFuser. 
MaskFuser also achieves competitive performance on the online \textit{official CARLA Leaderboard\footnote{\url{https://leaderboard.carla.org/leaderboard/}}}


\begin{figure*}[htbp]
    \centering
    \subfloat[city heavy traffic\label{subfig:city normal traffic}]{
        \includegraphics[width=0.24\textwidth]{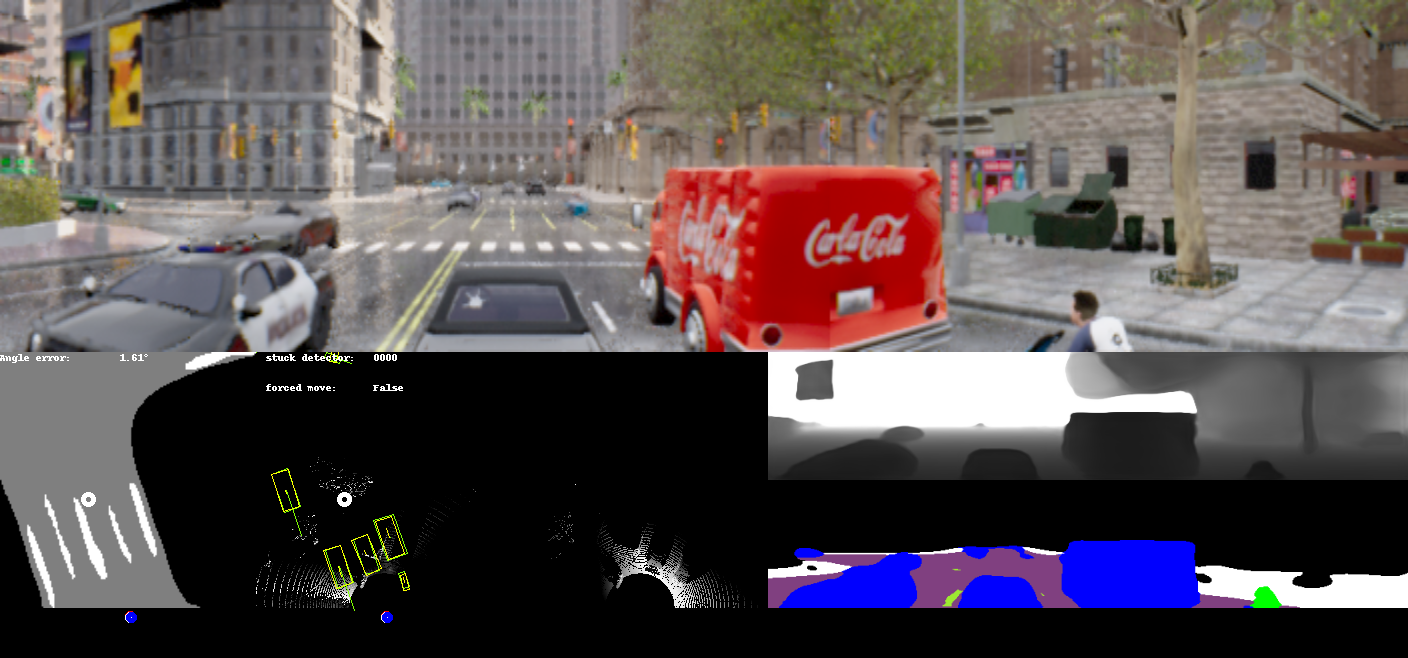}
    }
    \subfloat[city traffic light in distance\label{subfig:trafficlight in distance}]{
        \includegraphics[width=0.24\textwidth]{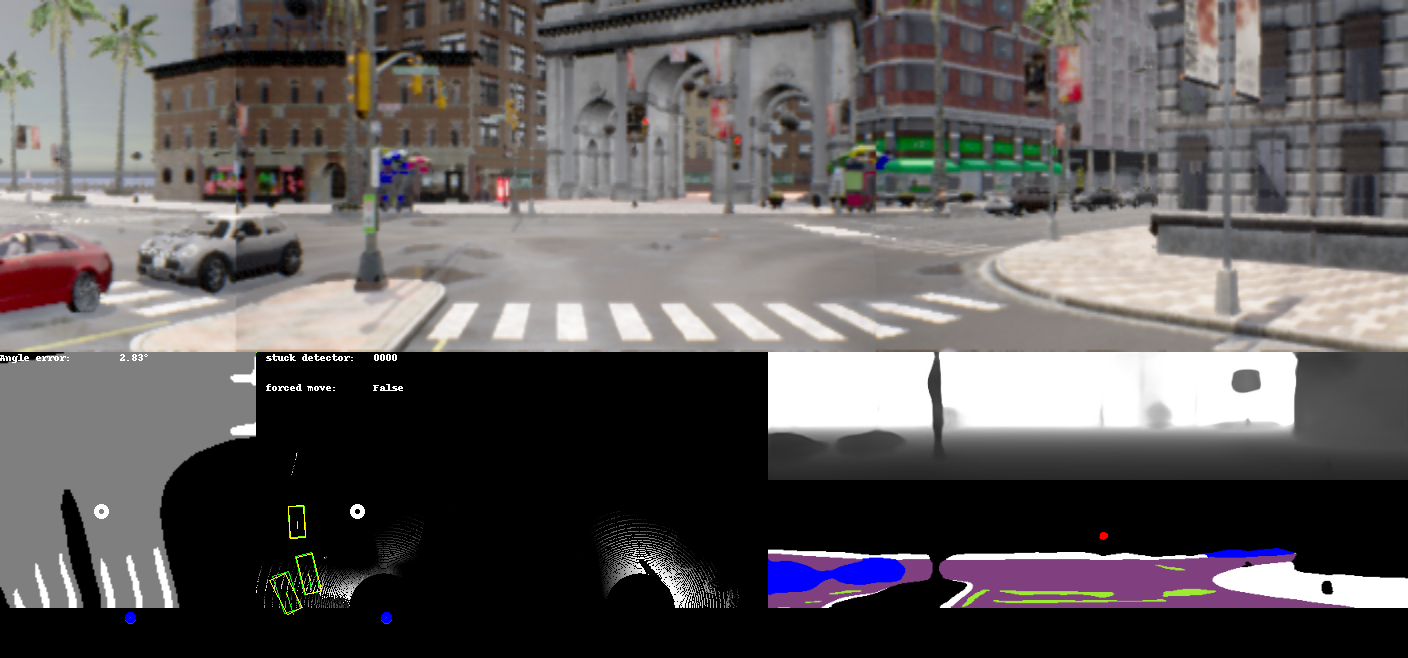}
    }
    \subfloat[city pedestrian ahead\label{subfig:city pedestrian ahead}]{
        \includegraphics[width=0.24\textwidth]{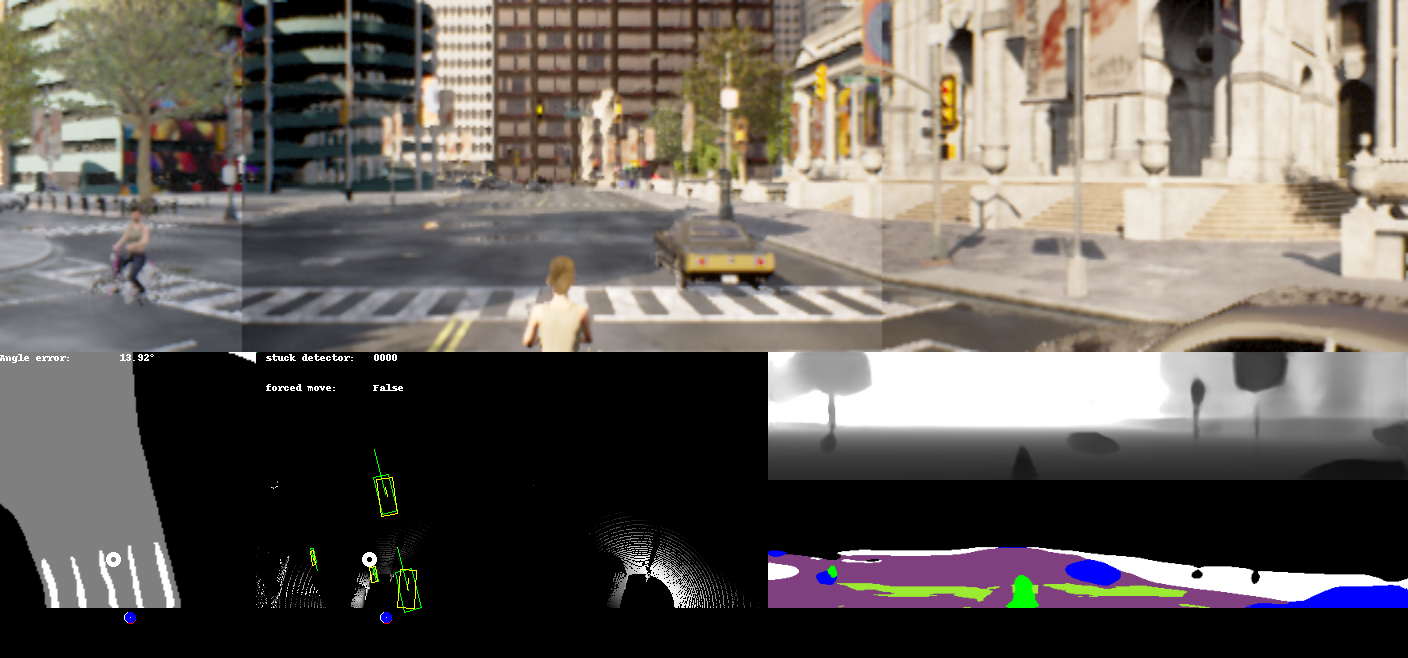}
    }
    \subfloat[city curve road\label{subfig:still}]{
        \includegraphics[width=0.24\textwidth]{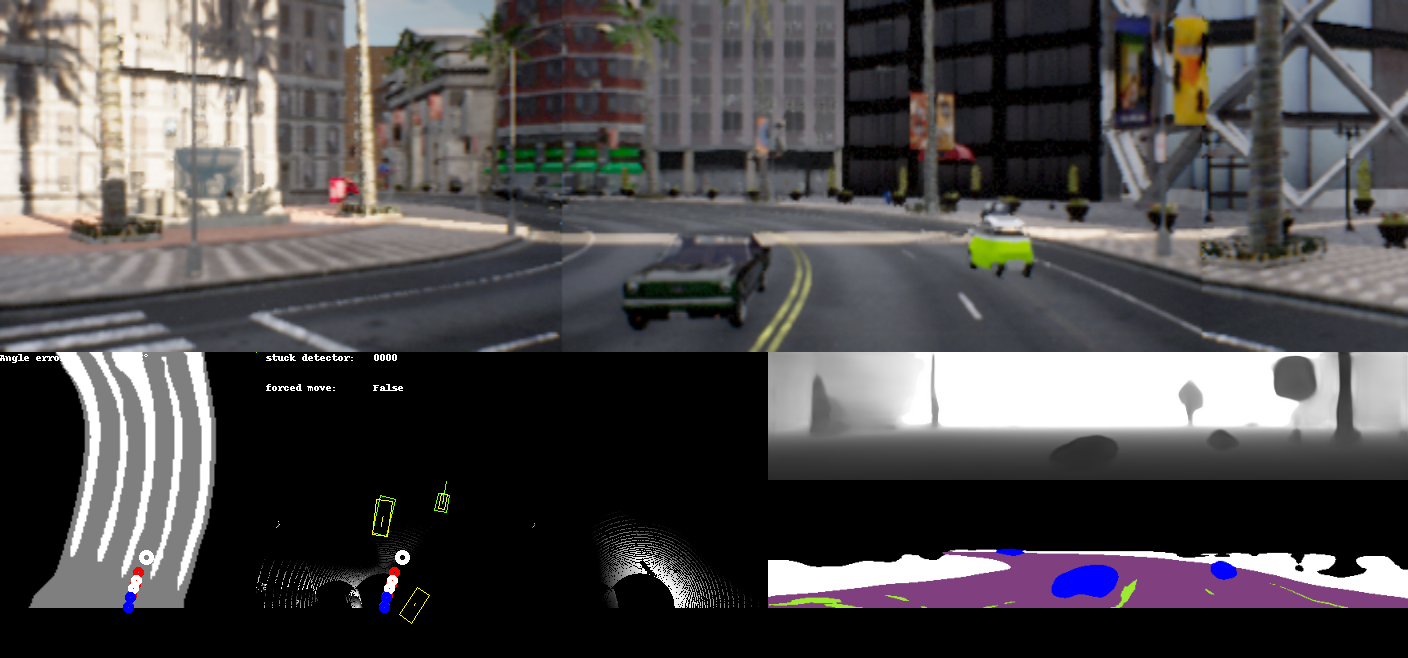}
    }
    
    \subfloat[city following the intersection\label{subfig:intersection}]{
        \includegraphics[width=0.24\textwidth]{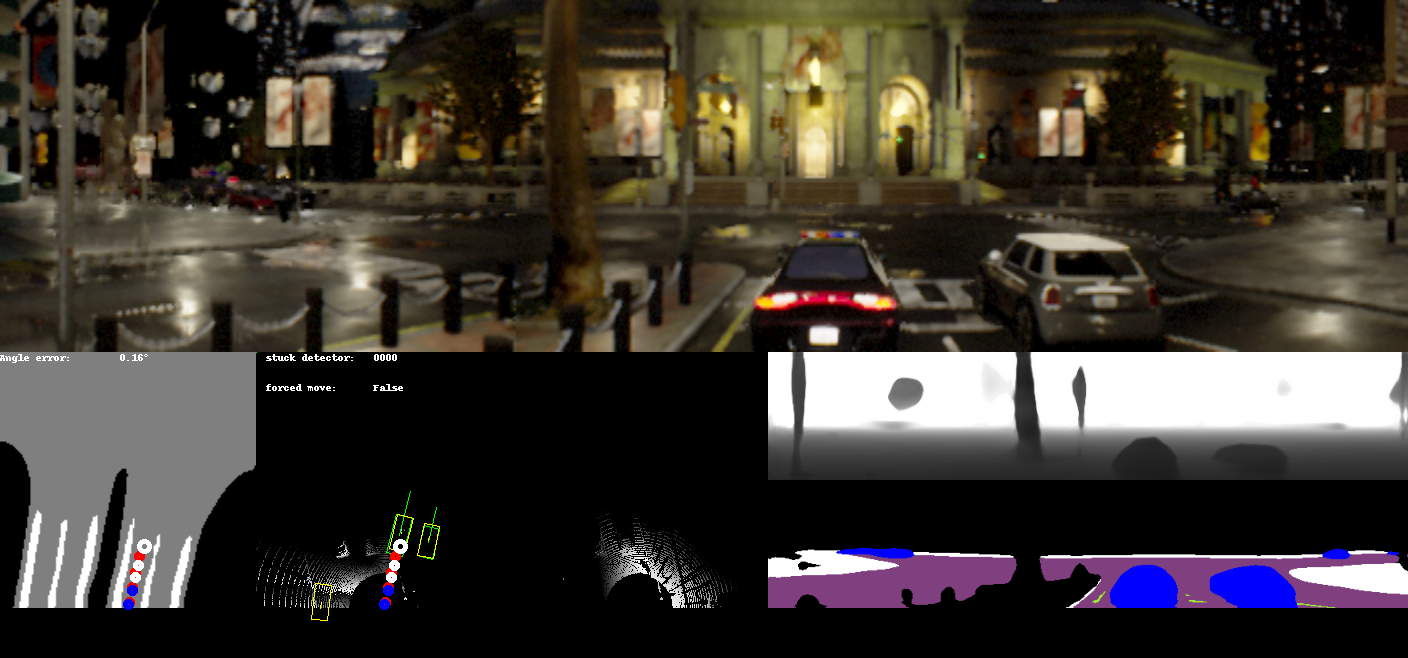}
    }
    \subfloat[city heavy rain on cross road\label{subfig:night_raining_cross}]{
        \includegraphics[width=0.24\textwidth]{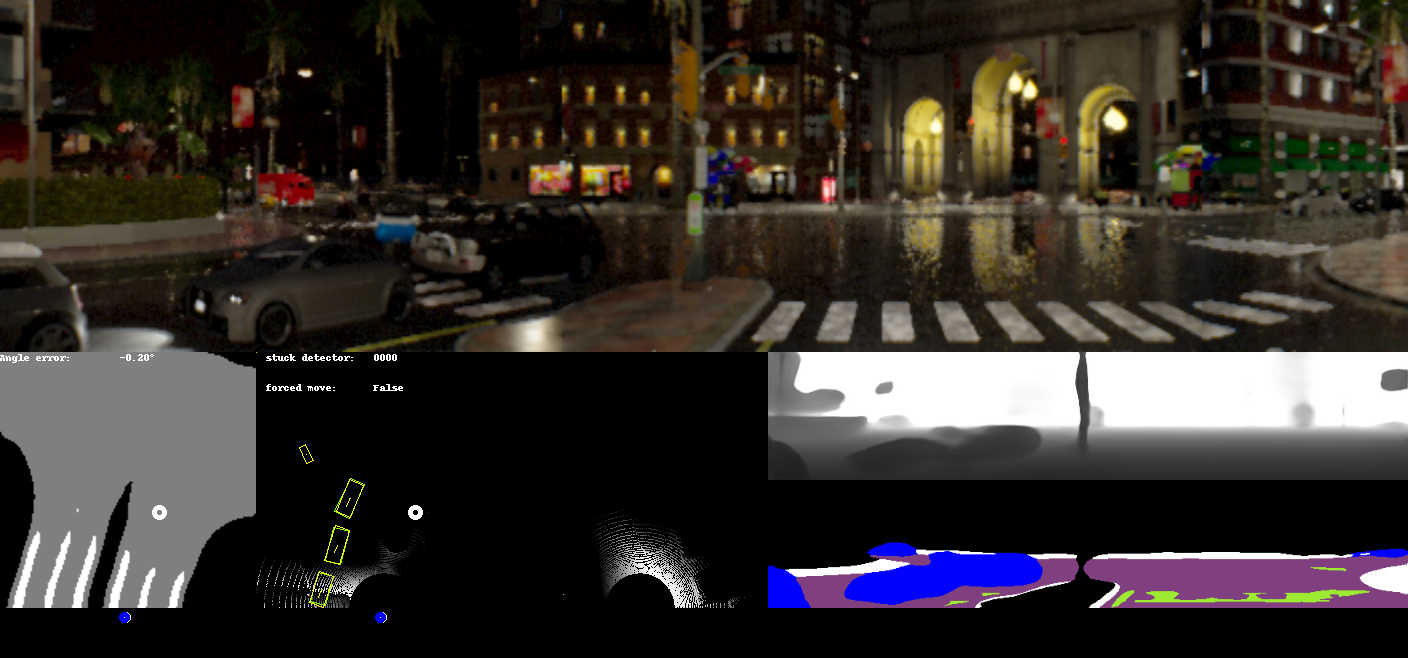}
    }
    \subfloat[city pedestrian ahead (night)\label{subfig:pedestrian}]{
        \includegraphics[width=0.24\textwidth]{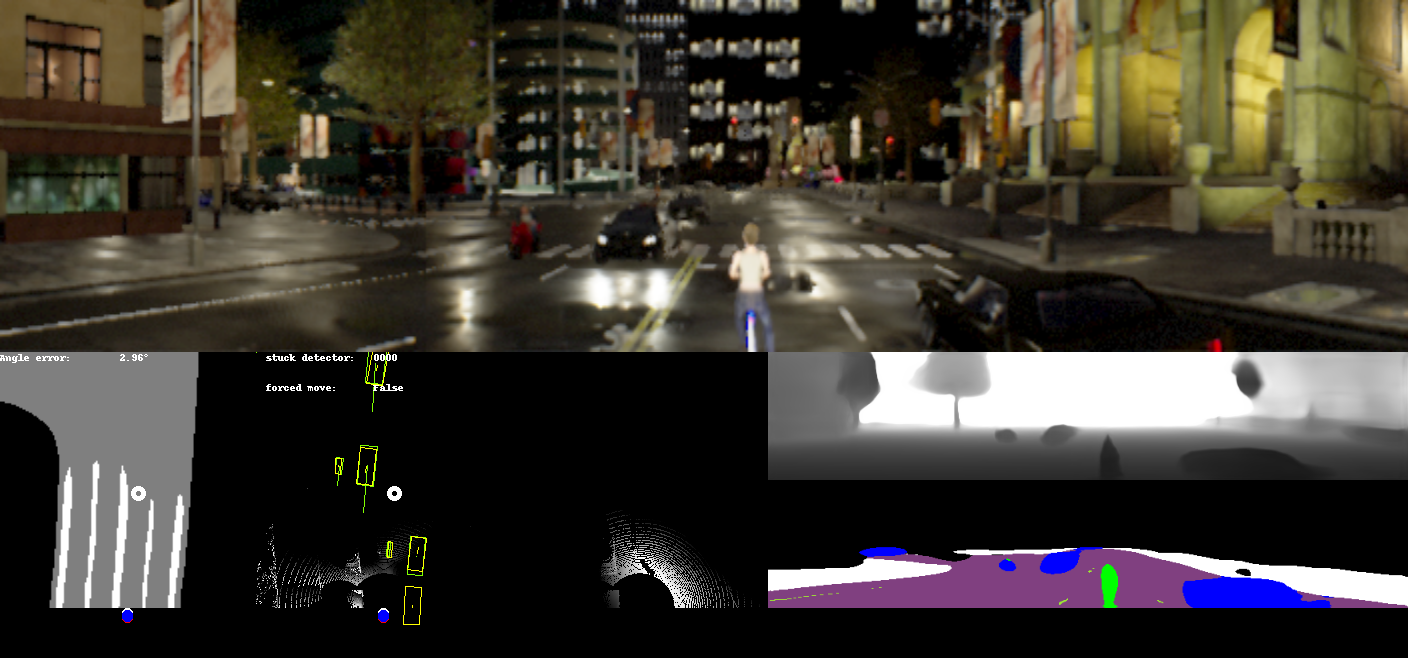}
    }
    \subfloat[country road (low light)\label{subfig:night_country_road_low_light}]{
        \includegraphics[width=0.24\textwidth]{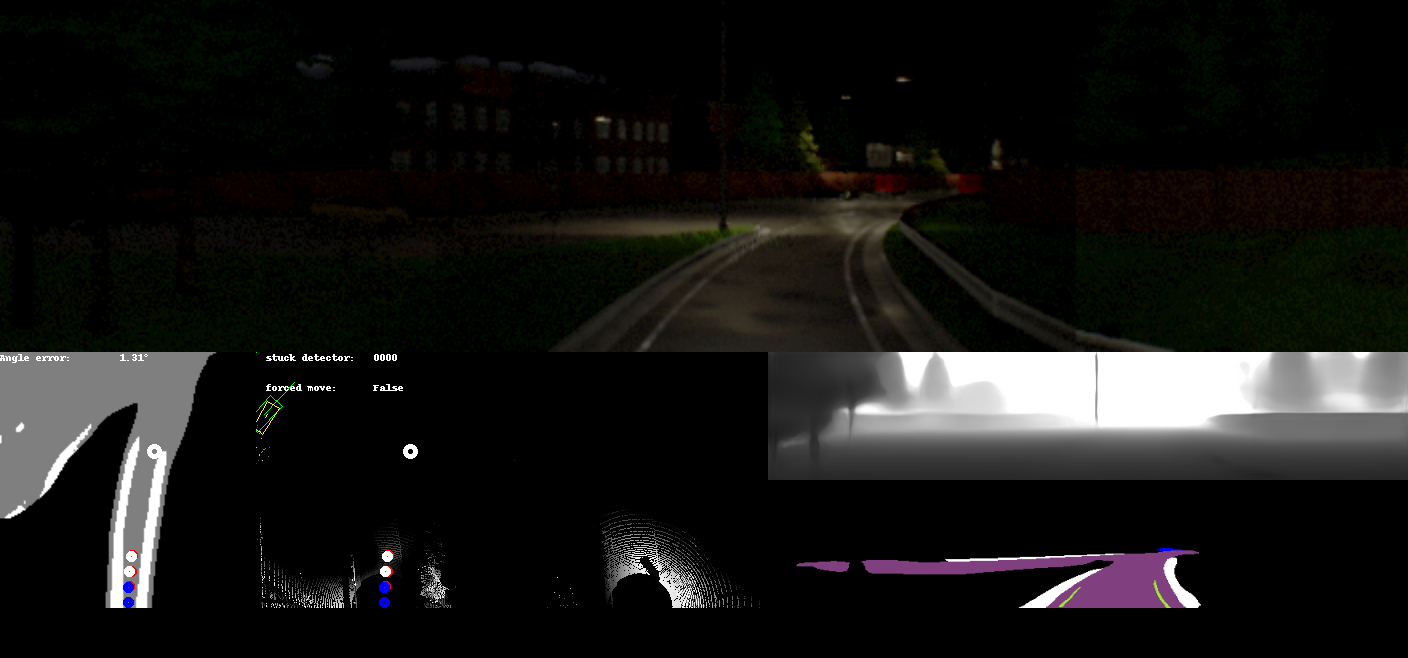}
    }

    \caption{Visualization of the driving process. The two rows list common driving environmental conditions during the day and night, such as pedestrian, heavy traffic, low light, and weather (raining) conditions. For segmentation map, legends are \textcolor[RGB]{0, 0, 0}{none}, \textcolor[RGB]{0,0,255}{vehicle}, \textcolor[RGB]{128, 64, 128}{road}, \textcolor[RGB]{255, 0, 0}{red light}, \textcolor[RGB]{255, 0, 0}{red light}, \textcolor[RGB]{157, 234, 50}{road line}, \textcolor[RGB]{0, 255, 0}{pedestrian}, and side walk (white).} 
    \label{fig:visdriving}    
\end{figure*}

\begin{figure}[hbpt]
    \centering
    \subfloat[Attention on curve road\label{fig:att_curve}]{
    \includegraphics[width=0.225\textwidth]{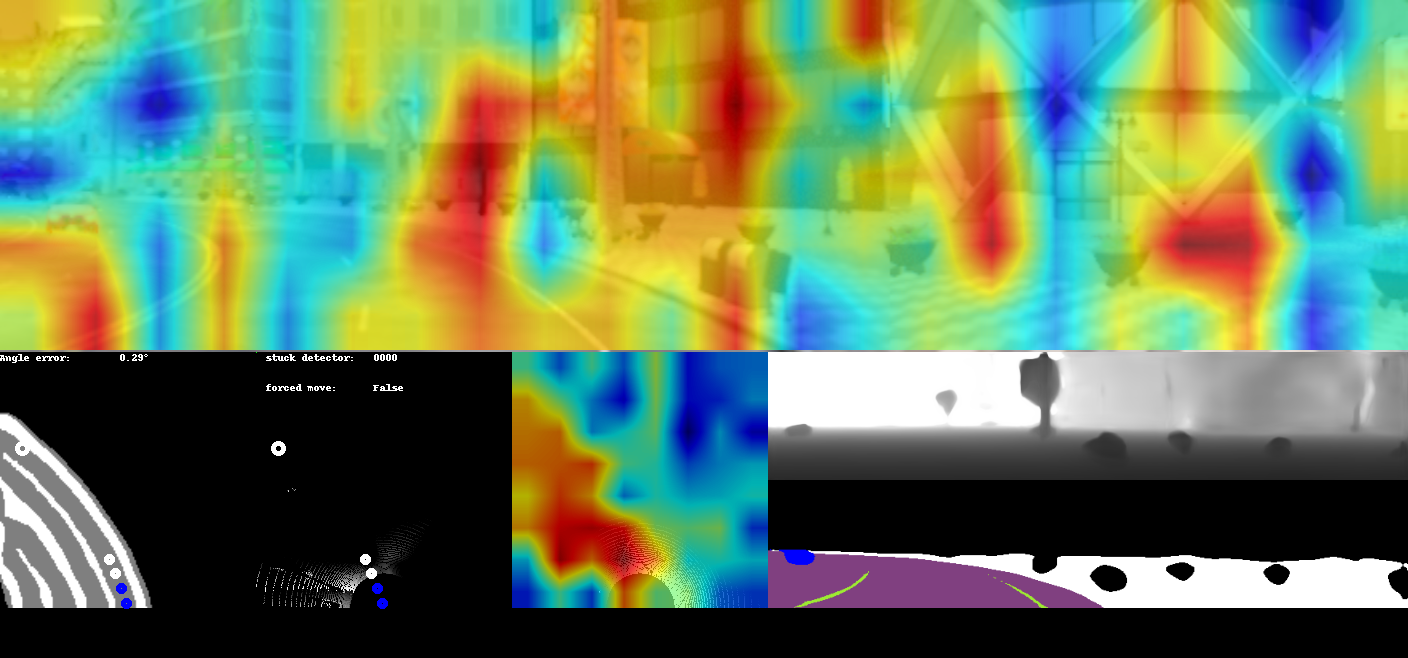}
    }
    \subfloat[Attention with red traffic light\label{fig:att_traffic_light}]{
    \includegraphics[width=0.225\textwidth]{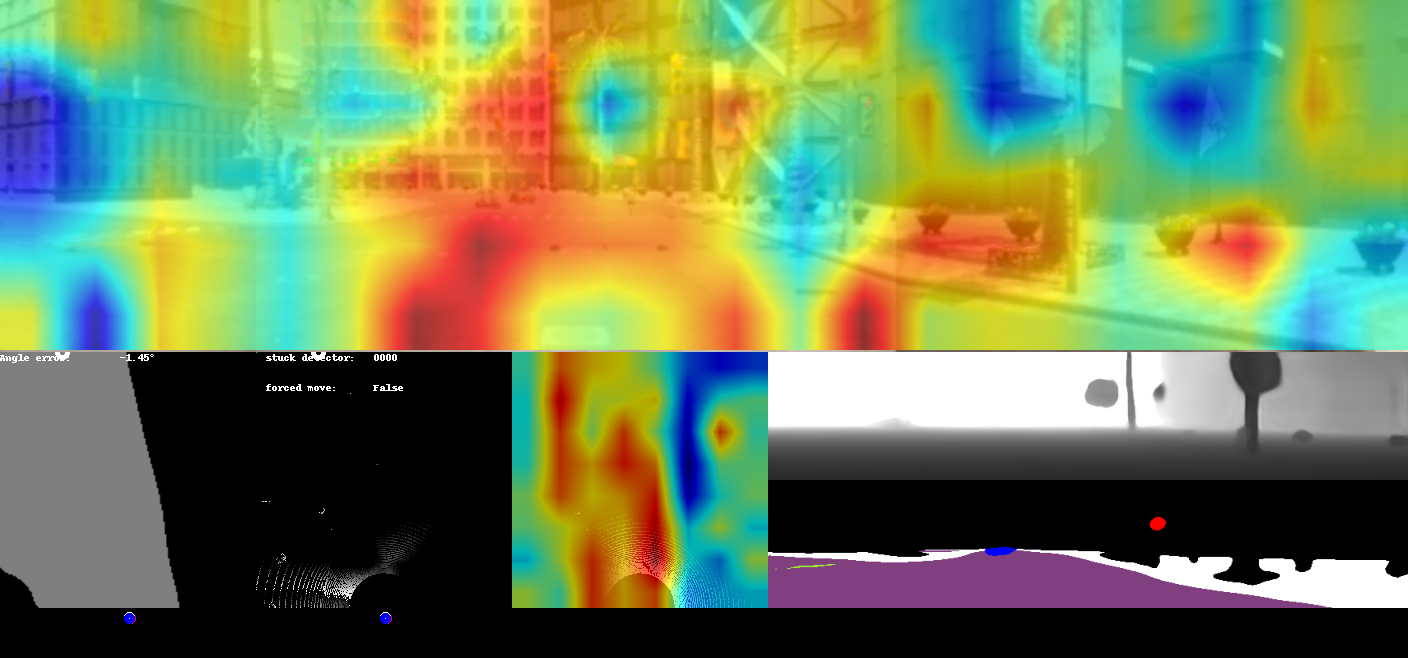}
    }
    
    \subfloat[Attention when following cars\label{fig:followingcars}]{
        \includegraphics[width=0.225\textwidth]{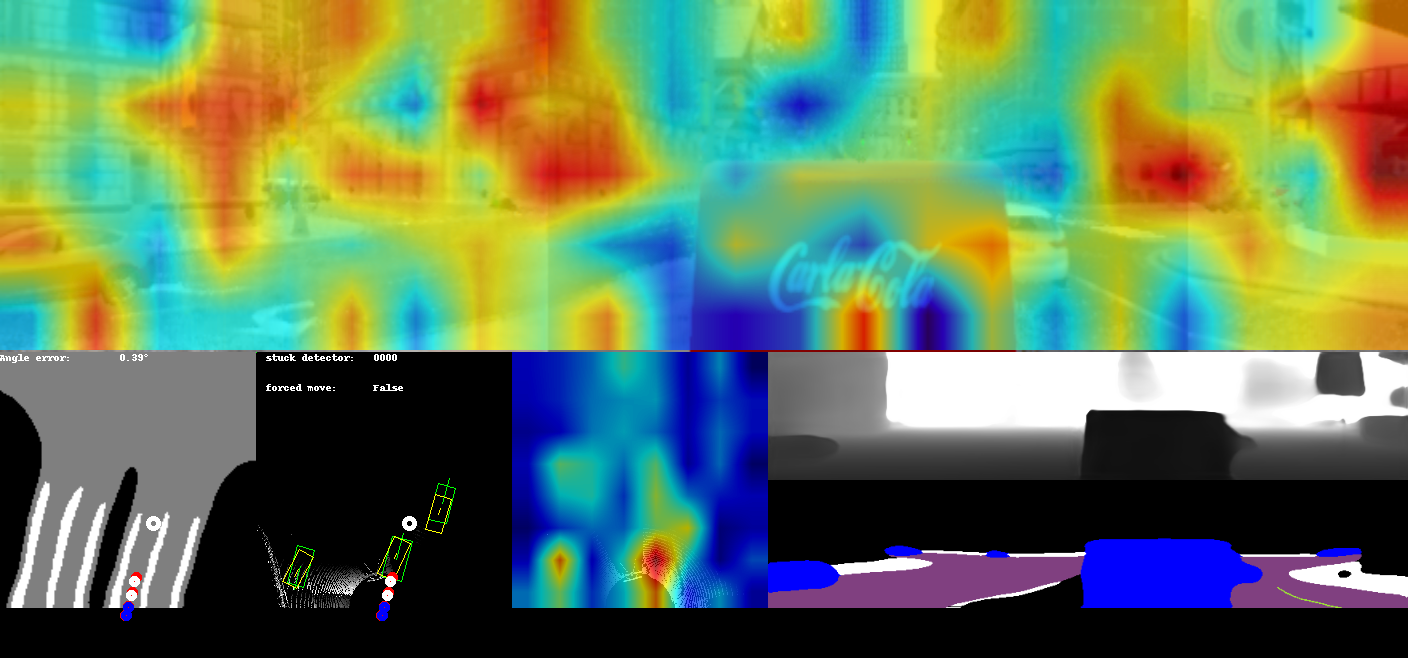}
    }
    \subfloat[Attention driving straight forward\label{fig:driving_straight}]{
        \includegraphics[width=0.225\textwidth]{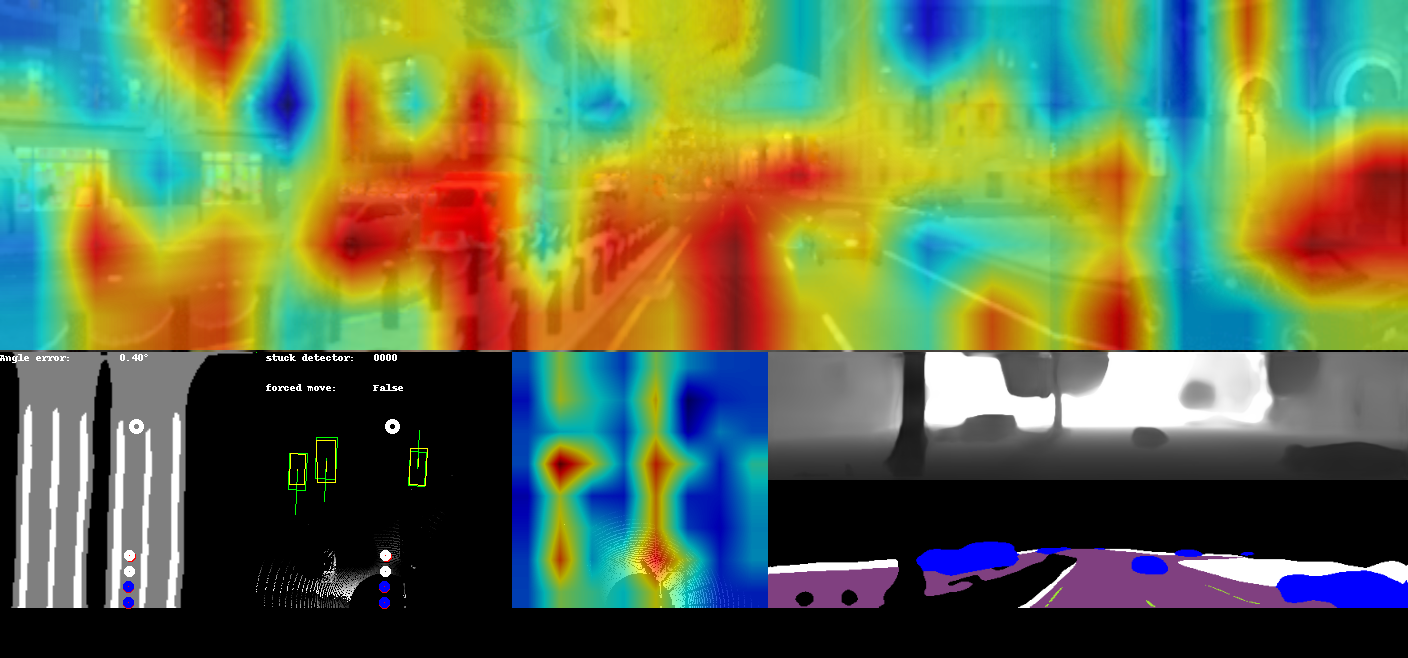}
    }
    
    \subfloat[Attention turning left on cross road\label{fig:let_turn}]{
        \includegraphics[width=0.225\textwidth]{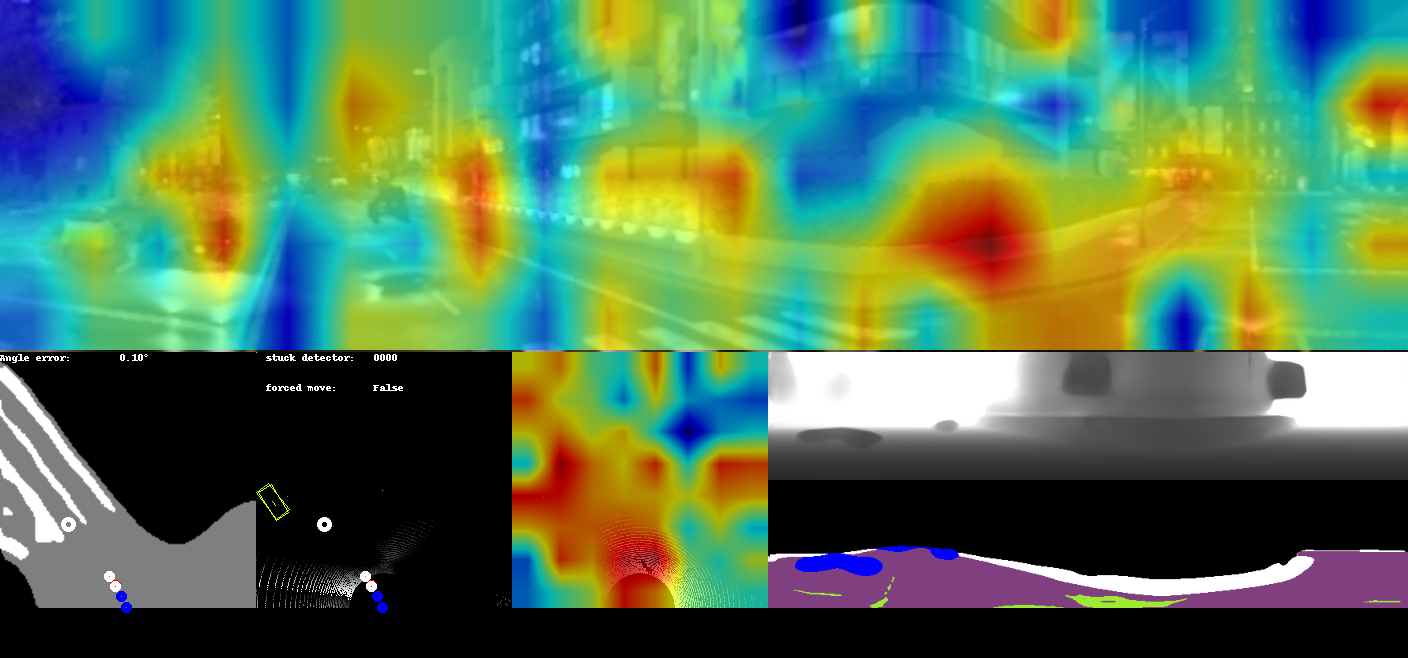}
    }
    \subfloat[Following Cars]{
    \includegraphics[width=0.225\textwidth]{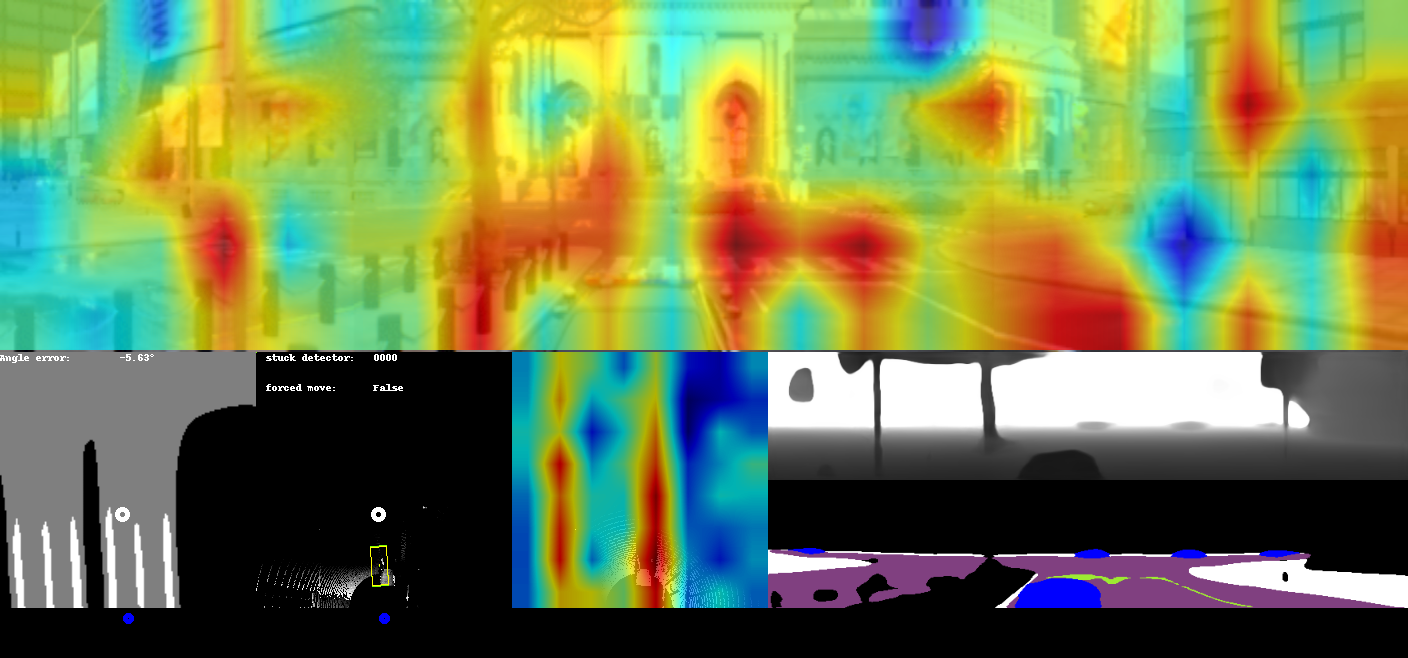}
    }
    
    \caption{Averaging attention visualization on joint tokens. The attentions are re-projected into range-view and BEV-view. We simultaneously report the BEV map prediction, depth prediction, and segmentation prediction with the attention maps.\label{fig:vis_attention}}
\end{figure}

\subsection{Visualization}\label{exp:visualization}

\textbf{Driving}
To give more intuitive explanations, we visualize the driving process by combining original sensory inputs with the predicted BEV map, segmentation map, and depth map in Fig.~\ref{fig:visdriving}. 
The solid perception results on common challenging scenarios suggest the efficiency of the proposed MaskFuser. 
Especially, the decent BEV map prediction suggests the efficiency of the proposed hybrid fusion and monotonic-to-BEV translation attention. 
  

\textbf{Attention Visualization}
The masked training provides better explainability on cross-modality driving. 
The results are shown in Fig.~\ref{fig:vis_attention}, where our methods show precise attention to both views, especially on the BEV view. It precisely locates objects, cars, lines, and potential barriers for making the decision.   
Here, we average attention from all attention heads to report a comprehensive attention map for driving. 
The results are shown in Fig.~\ref{fig:vis_attention} as listed. 
The attention results show considerable interpretability. 
As suggested in Fig.~\ref{fig:att_curve}, and Fig.~\ref{fig:let_turn}, when the car is turning on a curve road, the BEV attention expresses attention that is as oblique as the angle of the road curve corner. 
Also, Fig.~\ref{fig:att_traffic_light} denotes that the attention map could successfully give a highlight spot on the traffic light.
Fig.~\ref{fig:driving_straight} suggests an interesting straight attention line along the road direction. 
For all maps, the network normally pays more attention to 1) locations with more traffic conditions and 2) road conditions with their driving direction. 
Both points are quite rational and similar to what humans will pay attention to when driving. 
This supports the efficiency of the proposed network structure. 
Another observation is that the range view and BEV view attention some times respectively pay attention to different parts. 
Yet these parts are complementary to each other. 
This phenomenon also suggests the efficiency of the proposed joint representation.

\subsection{Driving with Partially Masked Sensors}\label{exp:mae}
\begin{figure}[t]
    \centering
    \subfloat[partial car in distance\label{subfig:farcar}]{
        \includegraphics[width=0.22\textwidth]{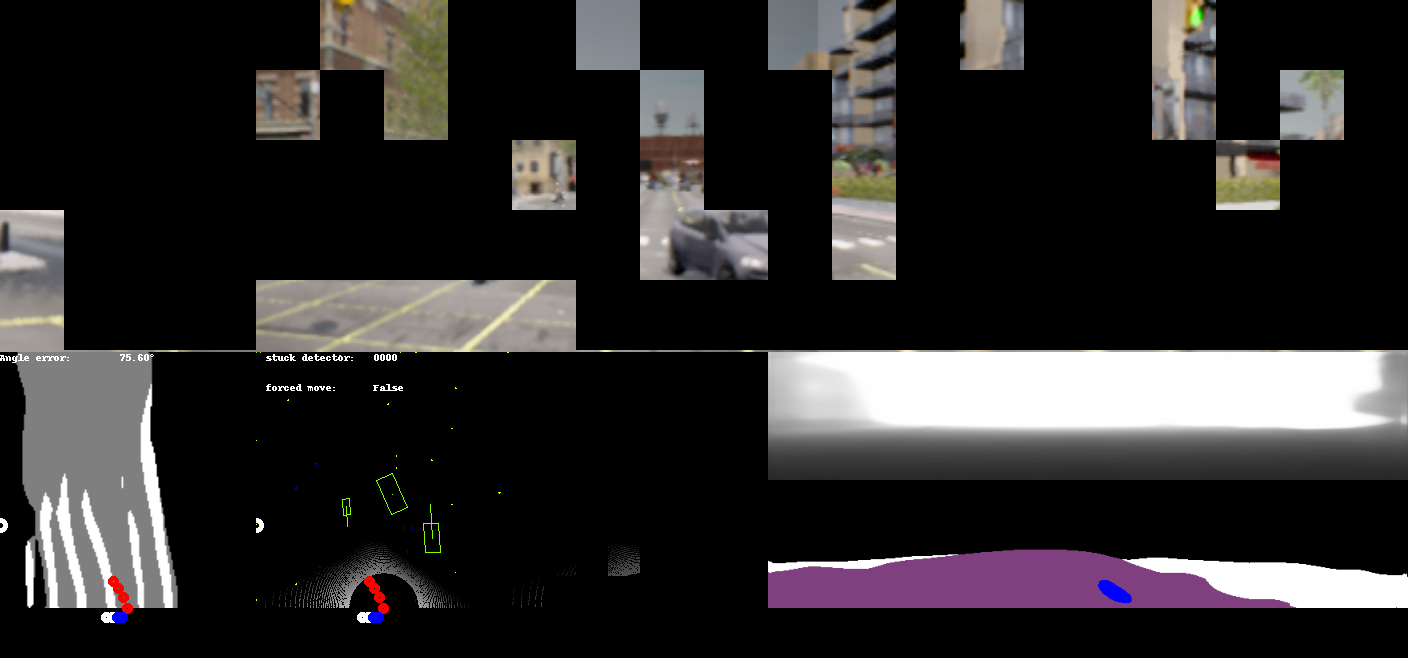}
    }
    \subfloat[partial car nearby\label{subfig:nearcar}]{
        \includegraphics[width=0.22\textwidth]{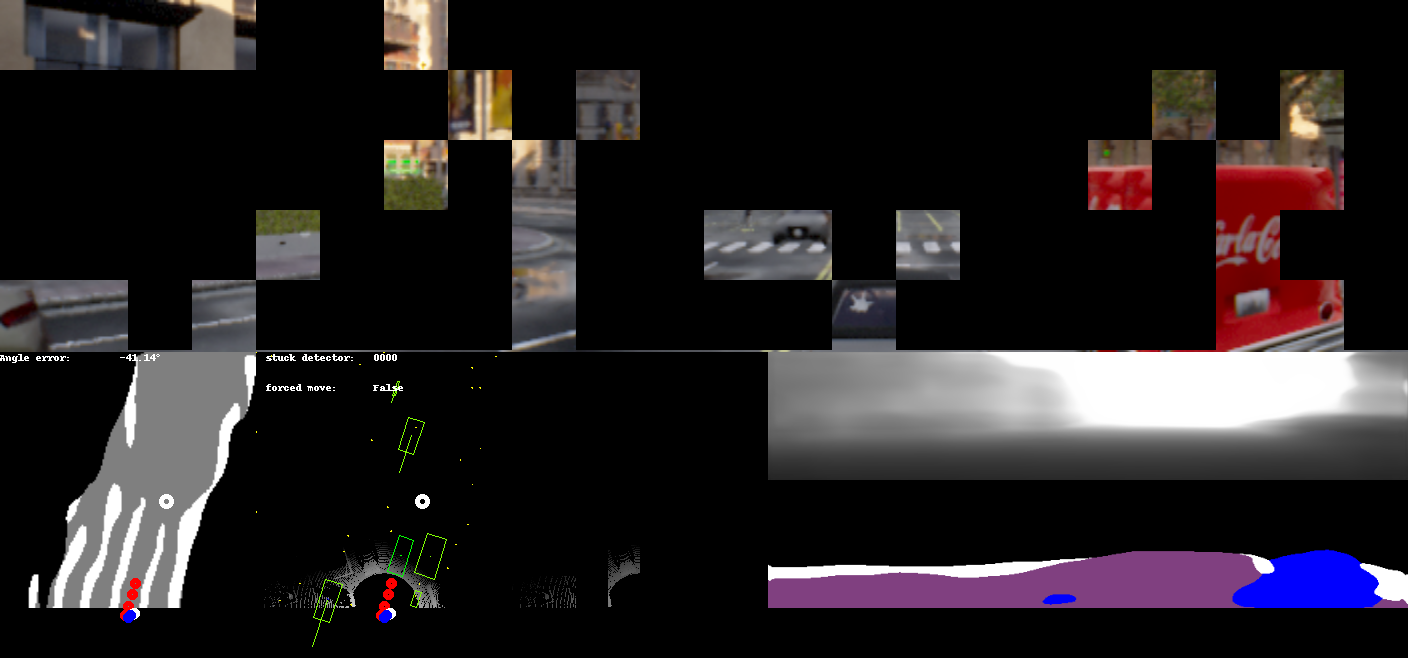}
    }
    \caption{The driving process with masked ratio $75\%$}
    \label{fig:drivingvis}    
\end{figure}

As mentioned in the introduction, current methods are under the assumption that the driving is with high-quality stable sensory inputs.
Yet, none of the previous works have explored improving driving performance by considering \textit{sensory broken}~\footnote{Such as cameras or LiDAR sensor is partially covered by mud or water during driving.} problem, which is of vital importance for real-life safety concerns.
We conduct sensory-damaged driving experiments by partially masking both image and LiDAR input with masking ratio $r\%$.
The driving processes are visualized in Fig.~\ref{fig:drivingvis}.
The damaged sensory inputs severely impact the depth map, which also leads to distortion on the far side of the BEV map. 
However, visualization suggests that MaskFuser could still successfully predict the semantic map and BEV map with rational \textit{guess} of the perception elements even if $75\%$ of inputs are invisible. 
In Fig.~\ref{fig:drivingvis}, MaskFuser successfully recognized the car in the segmentation map given visible fragments.
This supports our analysis (Sec.~\ref{subsec:cmmae}) that MaskFuser could largely increase the driving stability given damaged sensory inputs. 

\begin{table}[t]
\setlength\tabcolsep{5pt}
\caption{\label{tb:maskedfusion} Evaluation on LongSet6 by damaged sensory, where mask ratio $r$ denotes the partition of sensory features, is not visible to the encoder, and Star $^\star$ denotes fine-tuned with masked data augmentation.}
\centering
\resizebox{0.98\columnwidth}{!}{
\begin{tabular}{clcccc}
\toprule
\multicolumn{1}{l}{Mask $r$} & Methods    & LiDAR & DS $\uparrow$& RC$\uparrow$ & IS $\uparrow$ \\ \midrule
\multirow{5}{*}{75$\%$}        
                           & NEAT$^\star$        &  $\times$     & 2.16   & 6.92   &  0.03  \\
                           & TransFuser$^\star$  &  \checkmark &  5.08 & 10.78 &   0.05 \\
                           & MaskFuser  &  \checkmark  &  \textbf{6.65} & \textbf{13.93} &  \textbf{0.07}  \\ \hline
\multirow{5}{*}{50$\%$}    & NEAT$^\star$        &   $\times$    & 5.64 & 18.15   & 0.14  \\
                           & TransFuser$^\star$  & \checkmark & 11.08   & \textbf{27.01} &  0.10  \\
                           & MaskFuser  &  \checkmark & \textbf{12.61}   & 25.15   &  \textbf{0.15}  \\ \midrule
\multirow{5}{*}{25$\%$}    
                           & NEAT$^\star$       &  $\times$  & 12.47 & 35.43   & 0.23   \\
                           & TransFuser$^\star$  & \checkmark & 23.49   & 58.17  & \textbf{0.33}  \\
                           & MaskFuser  &  \checkmark & \textbf{30.04}  & \textbf{64.15}   & 0.29   \\ \midrule
\multirow{5}{*}{0$\%$} 
 & NEAT      &   $\times$    &   20.63 & 45.31  &   0.54\\
& TransFuser   &   \checkmark   & {46.95} &     {89.64} & 0.52 \\
& MaskFuser     &  \checkmark  &  {49.05}   &  {92.85}  &   0.56      \\ \hline
\rowcolor{Gray8} & Expert     &  NA &  75.83  &   89.82  &  0.85     \\
                           
                           \bottomrule
\end{tabular}
}
\end{table}


To further illustrate the stability of MaskFuser, we report driving performance under different masked ratios in Table~\ref{tb:maskedfusion}.
For a fair comparison, we fine-tune the baseline models by applying masking data-augmentation operation with supervised data. 
RL-based models are unstable after masking. Thus we just conduct damaged experiments with fusion-imitation-based baselines for a fair comparison. 
MaskFuser outperforms ($\Delta$) all previous best results by $6.55~(27.8\%)$, $1.53~(13.8\%)$, $1.57~(30.9\%)$, respectively given sensory masking ratio $25\%$, $50\%$, and $75\%$.
It is rational in twofold 1) MaskFuser fuse multi-modality inputs, where the two modalities can complement each other to alleviate the negative effect brought by sensory damage 2)  MaskFuser utilizes joint tokens as representation, where the share encoding and masked reconstruction (Sec.~\ref{subsec:cmmae}) enrich each semantic token with surrounding features.

\subsection{Ablation Study on Pre-Training }

\begin{figure}[hbpt]
    \centering
    \subfloat[Reconstruction Loss\label{subfig:rec_token}]{
        \includegraphics[width=0.225\textwidth]{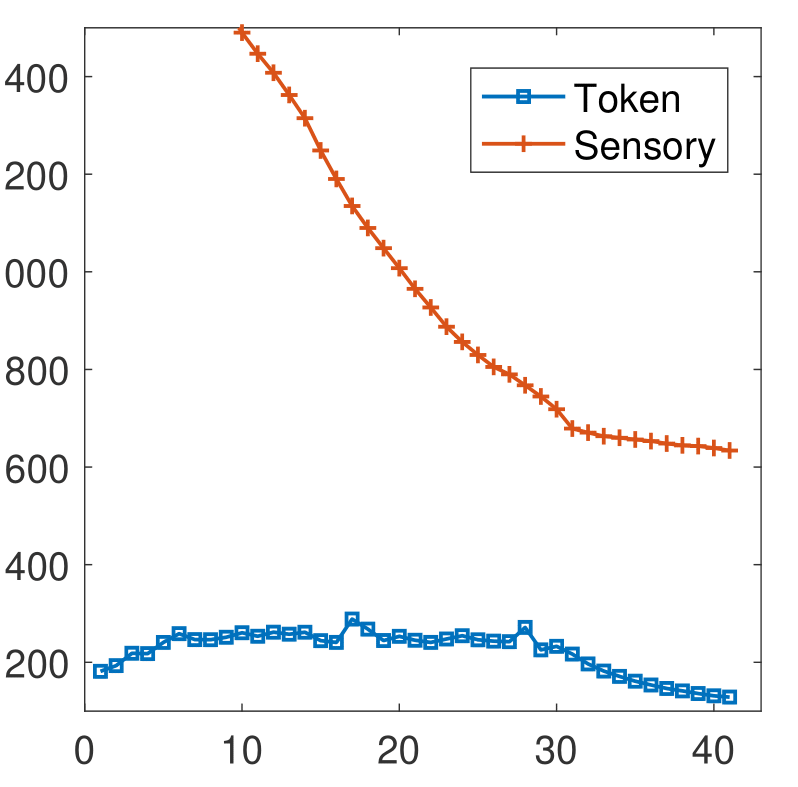}
    }
    \subfloat[Segment Loss\label{subfig:seg_token}]{
        \includegraphics[width=0.225\textwidth]{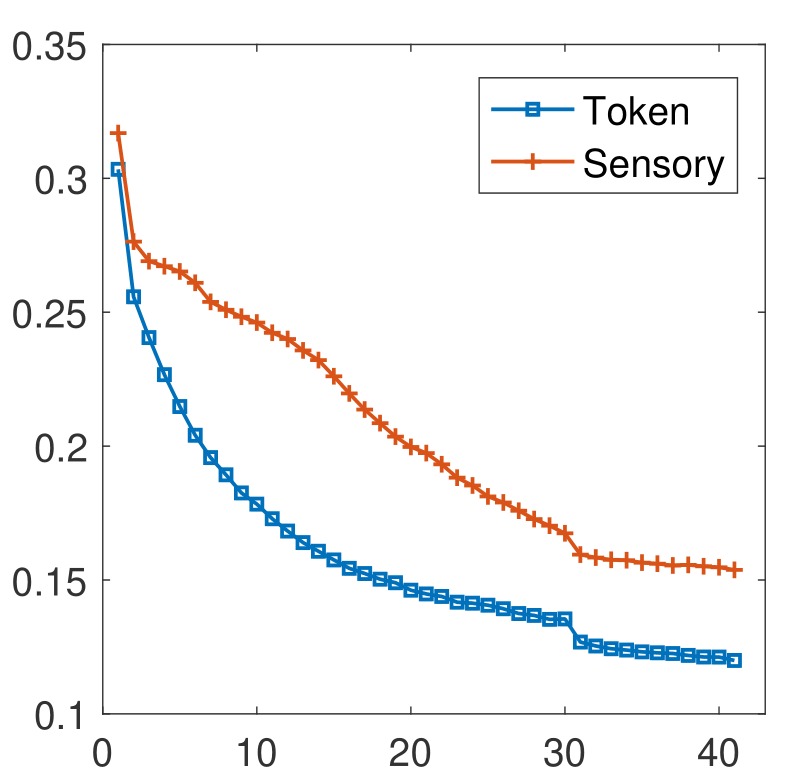}
    }
    
    \subfloat[BEV Loss\label{subfig:bev_token}]{
        \includegraphics[width=0.225\textwidth]{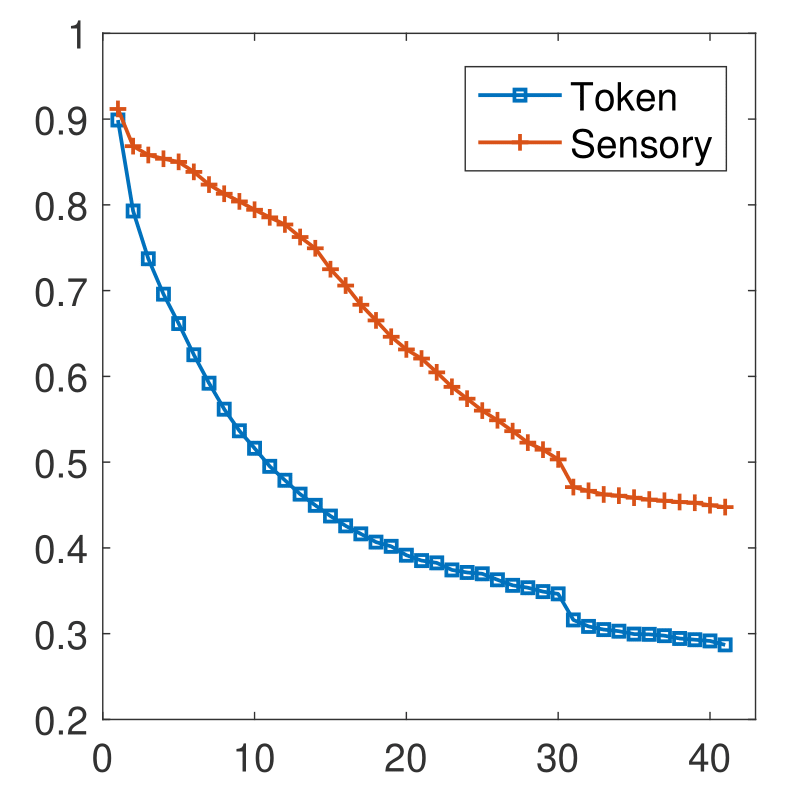}
    }
    \subfloat[Depth Loss\label{subfig:depth_token}]{
        \includegraphics[width=0.225\textwidth]{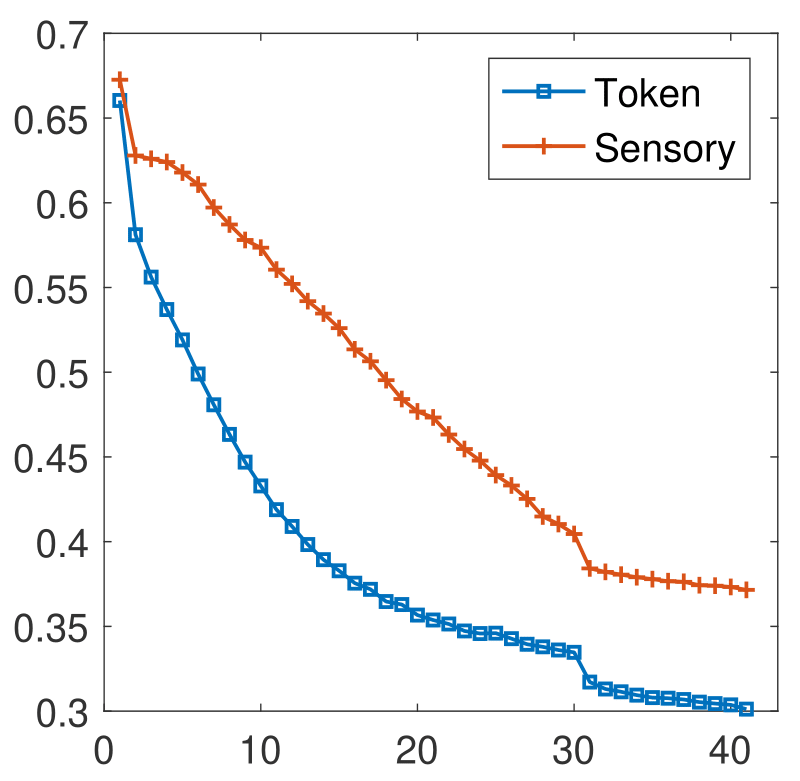}
    }
    \caption{Convergence comparison on different construction targets. The orange curve denotes reconstruction on original sensory inputs while the blue curve denotes it on the token itself. The auxiliary losses are achieved under masked ratio $75\%$.}
    \label{fig:ablationrec}
\end{figure}

\boldparagraph{Reconstruction Token vs. Reconstruction Sensory Inputs}\label{ap:mbtab}
Since the masking operation is applied to the semantic tokens, we have two strategies to realize the masked reconstruction training.
1) The target is to reconstruct the masked semantic tokens themselves. 
2) The target is to reconstruct original sensory inputs. 
We conduct an ablation study on both reconstruction strategies and compare the reconstruction losses in Fig.~\ref{fig:ablationrec}.
We simultaneously report the reconstruction loss and the auxiliary losses.
The reconstruction loss in Fig~\ref{subfig:rec_token} denotes that directly reconstructing original sensory inputs given unified token representation have a more clear descending trend compared to reconstructing the token itself. 
For reconstruction of the token itself, the loss will first rise and then descend. 
This is because the token represents itself and is also changing dramatically at early training stages. 
For auxiliary losses, both methods could achieve could convergence abilities. 
However, for utilizing the pre-trained weights, token-wise reconstruction does not suggest a clear improvement in the driving performance metrics. 
We suspect that reconstructing original sensory inputs could better help the encoder to 1) maintain rich perception details and 2) a harder reconstruction task may bring more supervision signals. 
In that case, we select to reconstruct the original sensory inputs in our pretraining stage. 

\subsection{Ablation Study on Performance}\label{exp:ablation}
\begin{figure}[hbpt]
    \centering
    \subfloat[Waypoints Loss\label{subfig:wp}]{
        \includegraphics[width=0.22\textwidth,height=2.8cm]{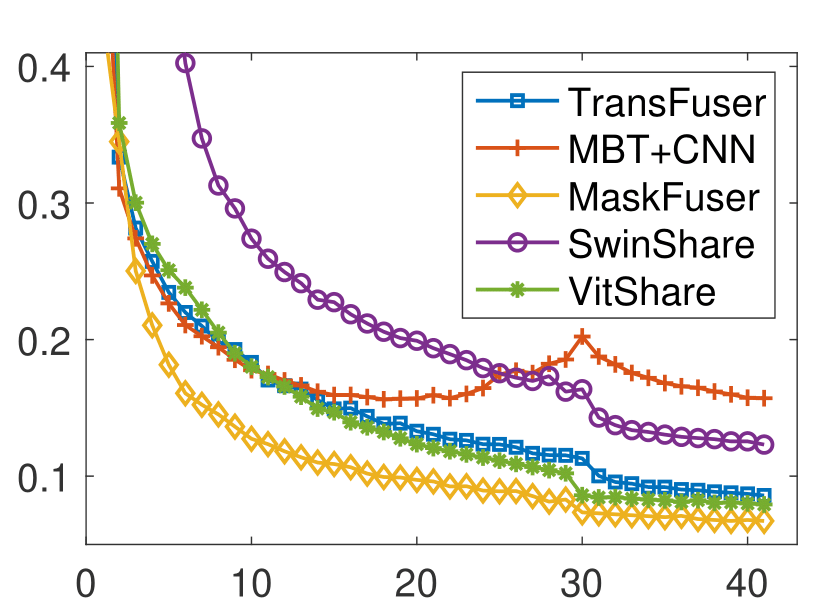}
    }
    \subfloat[Segment Loss\label{subfig:seg}]{
        \includegraphics[width=0.22\textwidth,height=2.8cm]{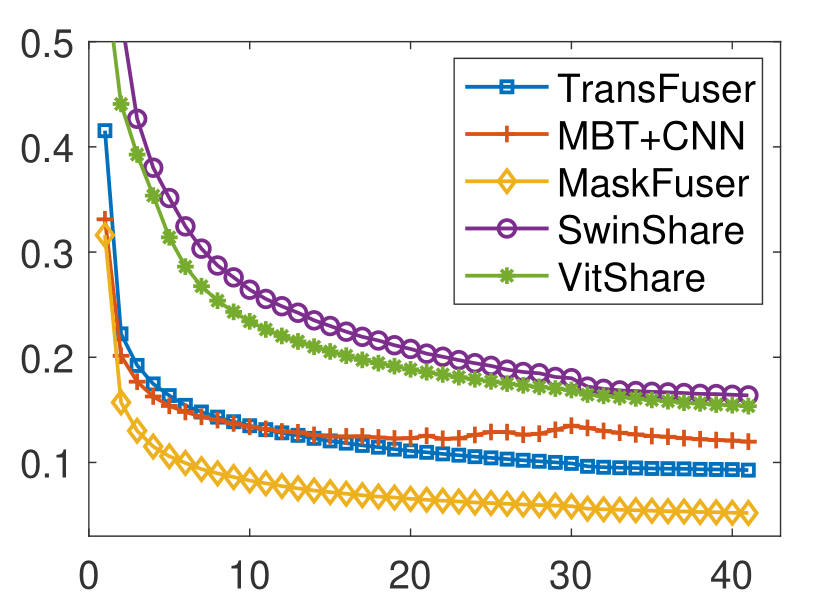}
    }

    \subfloat[BEV Loss\label{subfig:bev}]{
        \includegraphics[width=0.22\textwidth,height=2.8cm]{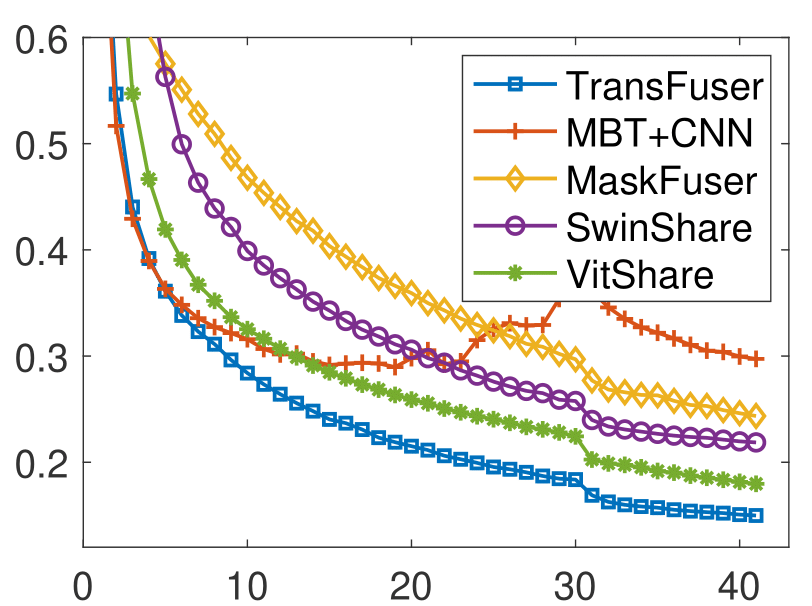}
    }
    \subfloat[Depth Loss\label{subfig:depth}]{
        \includegraphics[width=0.22\textwidth,height=2.8cm]{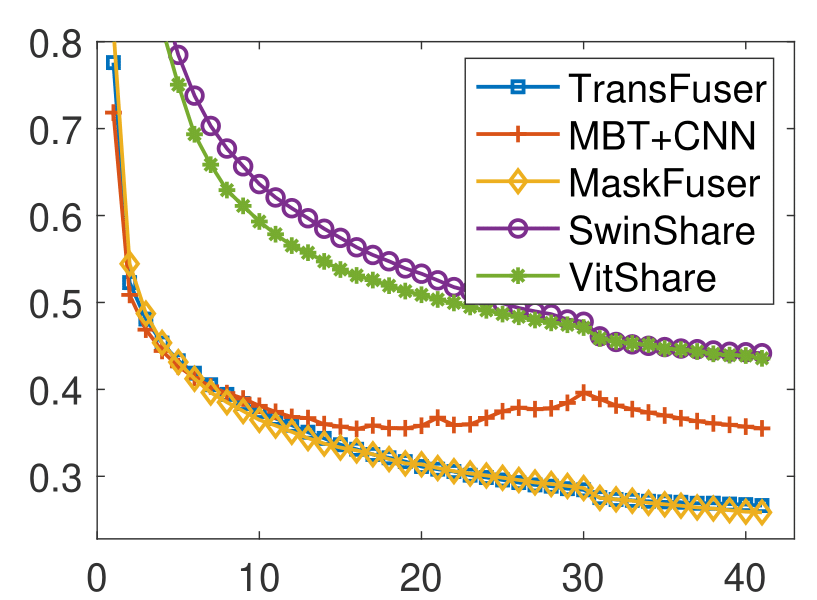}
    }
    \caption{Convergence comparison on loss curves, where x-axis and y-axis represent epochs and loss value respectively.}
    \label{fig:convergence}
\end{figure}

We respectively report the loss curve and the driving metrics of all comparison candidates in Fig.~\ref{fig:convergence} and Table~\ref{tb:ablation}.
We conduct discussions about several important architectural designs based on these comparisons.
\begin{table}[hbpt]
\setlength\tabcolsep{4.55pt}
\label{tb:ablation}
\caption{Ablation on LongSet6. Att. denotes the attention type for early fusion. Fusion denotes methods for late fusion.}
\centering
\resizebox{0.98\columnwidth}{!}{
\begin{tabular}{ccccccc}
\toprule
\multicolumn{1}{l}{}                                                   Att.   & Token & Share & Fusion & $\Delta$DS     & $\Delta$RC     & $\Delta$IS    \\ \midrule
 $\times$  &  $\times$  &  $\times$ & CNN    & -15.09 & -18.32 & -0.20 \\
                                                                         
                                        $\times$   &  \checkmark  & $\times$ & Swin   & -5.57  & -5.14  & -0.04 \\
                                                                         $\times$ &  \checkmark   & $\times$ & ViT    & -3.98  & -3.51  & -0.05 \\ \midrule
                                                                         Plain  & $\times$   &  $\times$ & CNN    & -5.43  & -4.97  & -0.04 \\
                                                                         Plain   &  \checkmark & $\times$ & Swin   & -4.92  & -4.41  & -0.03 \\
                                                                         Plain   & \checkmark  & $\times$  & ViT   & -4.15  & -4.09  & -0.02 \\
                                                                          Plain  &  \checkmark  &  \checkmark  & ViT    & -2.07  & -3.09  & +0.01 \\ \midrule
 MBT &   $\times$   & $\times$   & CNN    & -12.01 & -11.45 & -0.18 \\
                                                                        MBT    &  \checkmark &  $\times$  & Swin   & -2.01  & -3.14  & -0.01 \\
                                                                          MBT   &  \checkmark  & $\times$   & ViT    & -1.09  & -2.39  & +0.01  \\ \midrule
\rowcolor{Gray8} MBT   & \checkmark   & \checkmark  & ViT    & 48.07  & 92.03  & 0.55  \\ \bottomrule
\end{tabular}}
\end{table}

\noindent\textbf{Monotonic-to-BEV Attention:}
To illustrate the effect of MBT attention, we compare the model with three attention types, no attention ($\times$), plain attention through transformer layer (\textit{Plain}), and Monotonic-to-BEV Translation attention (\textit{MBT}), in Table~\ref{tb:ablation}. 
The MBT could increase the driving score by $0.98$, $2.14$ respectively on Swin and ViT base. 
However, the pure early fusion with the MBT+CNN combination would severely decrease the performance by 6.58, which suggests MBT attention between every CNN block are incompatible with independent branches. 
According to the convergence curves in Fig.~\ref{fig:convergence}, we find that MBT+CNN converges fast in top-3 at early training stages. However, the loss curve falls behind severely when training goes on. 
It is noted that the upper layers may have already been semantic features with compared weak spatial relations. 
As training goes on, continuously projecting semantic features also suppresses the comprehensive fusion of the driving context.
It supports our idea that pure geometric fusion is not perfectly suitable for comprehensive driving, which illustrates the efficiency of the proposed \textit{hybrid fusion} formation.


\noindent\textbf{Share Encoding:}
We compare the fusion strategy in the late fusion stage by whether the image and LiDAR stream merge into a shared encoder or remain in independent encoders. 
Since, Swin~\cite{liu2021swin} base and CNN base are not suitable flat token encoding, we only conduct this ablation on ViT base.  
Table~\ref{tb:ablation} denotes that letting various modalities share the transformer encoder could respectively boost the driving score from $0.85$ to $1.98$, which suggests a clear advantage. 
It is rational since the shared encoder aligns tokens from various modalities into a unified token space, which could bring deeper feature interaction thus enhancing the feature quality.

\noindent\textbf{ViT vs. Swin Transformer:}
According to Table~\ref{tb:ablation}, ViT outperforms Swin transformer by a solid margin ($+0.92$, $+2.08$, $+1.59$ respectively under different settings).  
It is rational since we actually using the transformer to perform late fusion on various modalities rather than only extracting range-view information. 
Swin transformer shares more similarity with CNN and ViT more similar to the language model. 
Considering the aim of late fusion is to conduct share encoding on joint representation, it is rational that ViT could outperform Swin by a solid margin. 

\subsection{Convergence Capability with Different Masking Ratio}
\begin{figure}[hbpt]
    \centering
    \subfloat[Waypoints Loss\label{subfig:rec_mask}]{
        \includegraphics[width=0.22\textwidth]{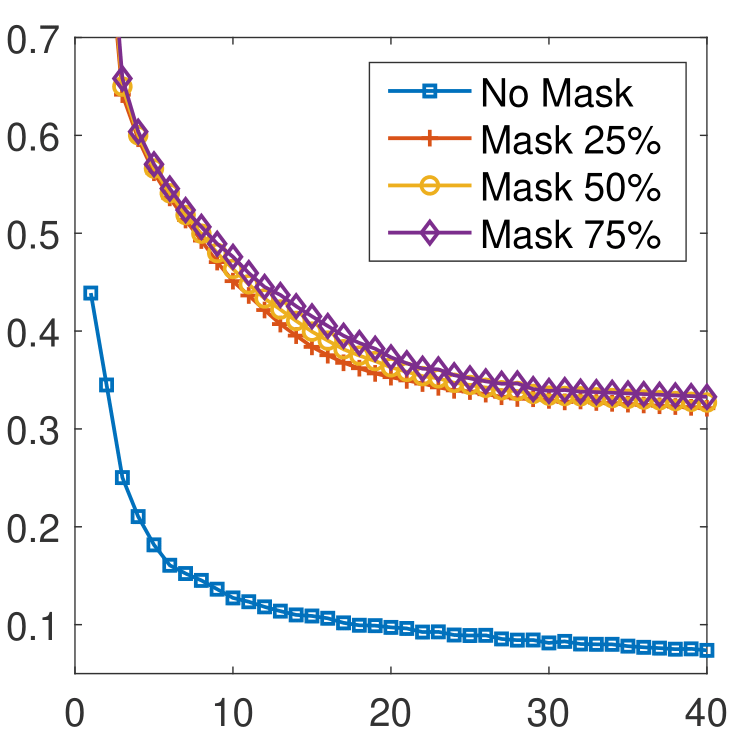}
    }
    \subfloat[Segment Loss\label{subfig:seg_mask}]{
        \includegraphics[width=0.22\textwidth]{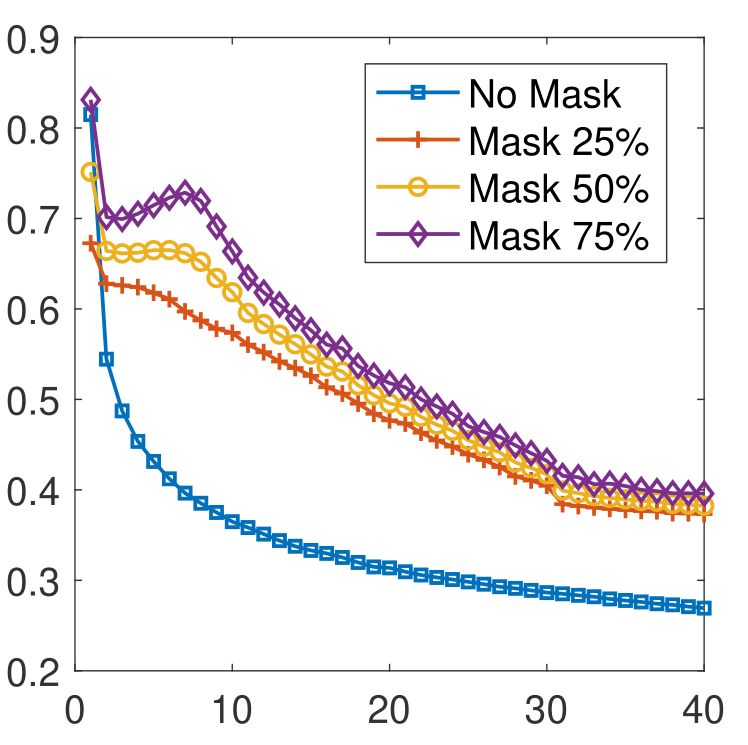}
    }
    
    \subfloat[BEV Loss\label{subfig:bev_mask}]{
        \includegraphics[width=0.22\textwidth]{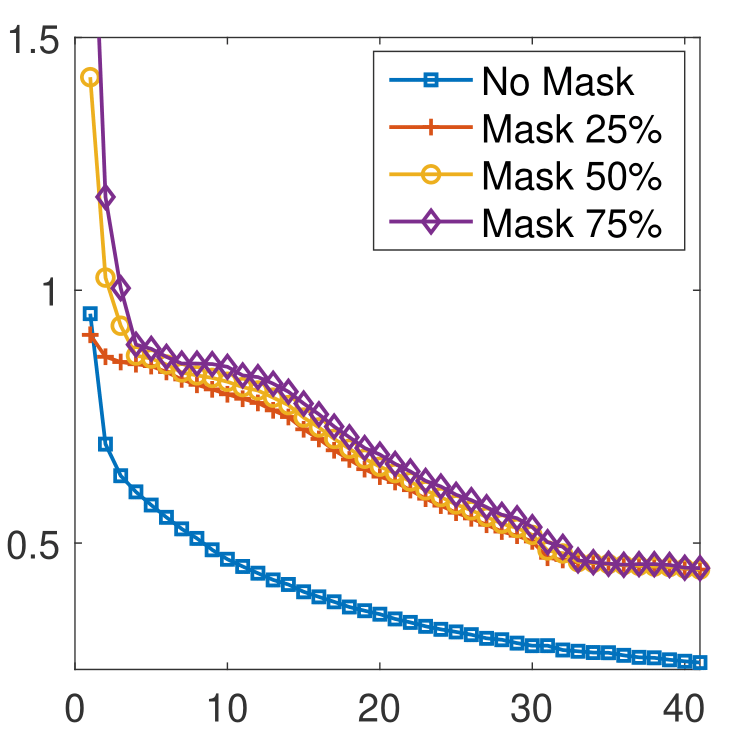}
    }
    \subfloat[Depth Loss\label{subfig:depth_mask}]{
        \includegraphics[width=0.22\textwidth]{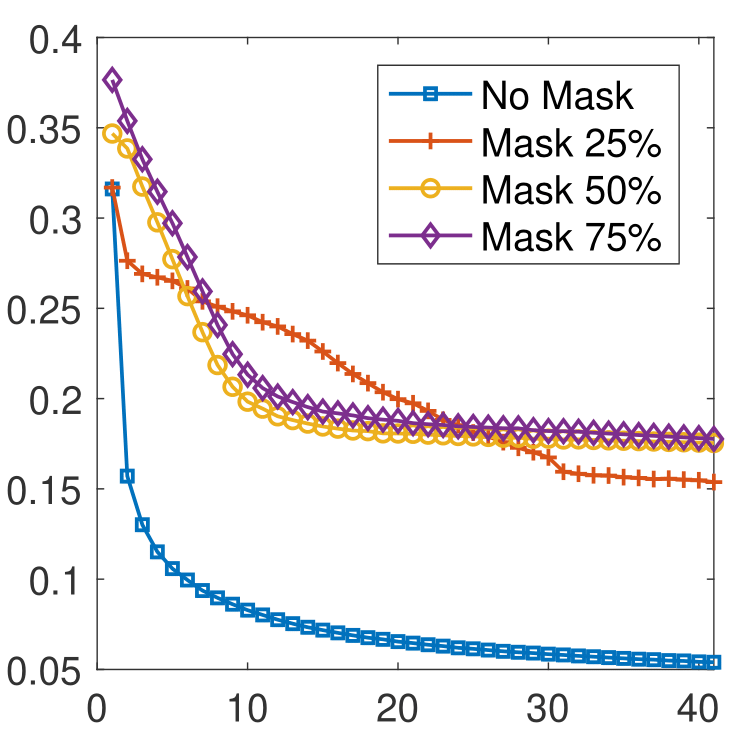}
    }
    \caption{Convergence comparison on different mask ratios.}
    \label{fig:ablation_mask_conv}
\end{figure}

We report the masked driving results in Table~\ref{tb:maskedfusion}
To understand more about how the perception results are affected by the masking ratio, we report more visualization results as well as the convergence properties of different masking ratios. 
The comparison of different masking ratios is reported in Fig.~\ref{fig:ablation_mask_conv}. 

It is observed that, for waypoints prediction and BEV map prediction, different mask ratios do not bring a clear difference in the loss curves. 
Although a lower masking ratio still suggests a lower loss, the gap is not as large as we expected. 
We suspect this is because BEV and waypoints prediction relies more on global information. 
For depth and segmentation map prediction, the lower mask ratio does have better convergence performance, which fits our expectations. 

\begin{figure}[hbpt]
    \centering
    \subfloat[]{
        \includegraphics[width=0.38\textwidth]{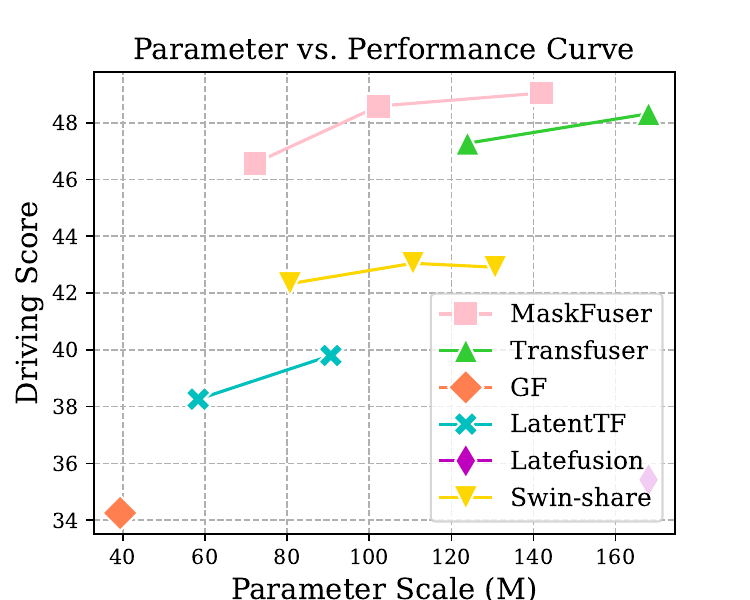}
        \label{subfig:speedpvp}
    }
    
    \subfloat[]{
        \includegraphics[width=0.38\textwidth]{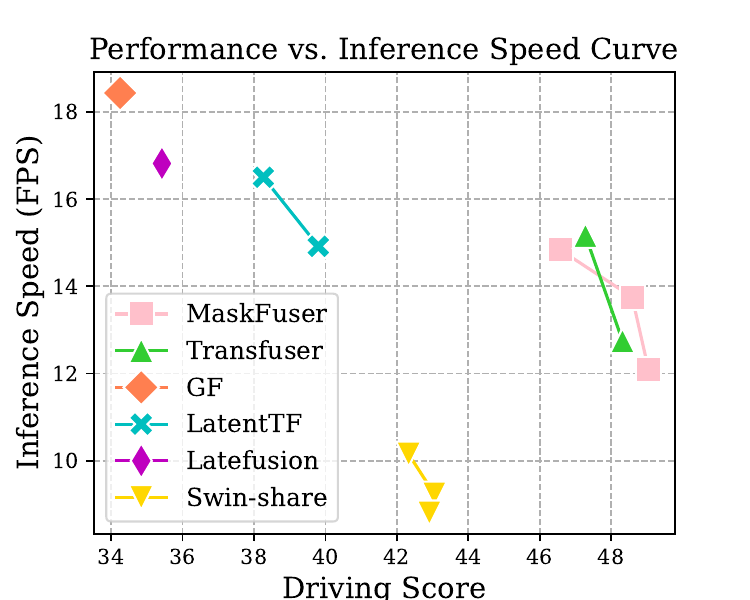}
        \label{subfig:speedfps}
    }
   \centering{\caption{Complexity vs. Performance curve of MaskFuser.}}
    \label{pvp}
\end{figure}

\subsection{Complexity.} 
Although MaskFuser has multiple decoders to train, only the HybridNet with a waypoints prediction module is required for inference. 
To illustrate the proposed methods are practical to use, we compare both the parameter scale and the inference time of MaskFuser with previous SOTA methods.
The curves are reported in Fig.~\ref{pvp}.
For the inference speed, MaskFuser is comparable with current SOTA TrasFuser~\cite{transfuser}.
It is noted that MaskFuser outperforms previous SOTA methods while using even fewer parameters. 



\section{Conclusion}
This paper presents MaskFuser, an innovative hybrid feature fusion framework designed for end-to-end autonomous driving systems. Distinguished as the first framework to utilize cross-modality (visual-LiDAR) masked training within a tokenized driving context, MaskFuser marks a significant advancement in the field. Extensive testing in the CARLA simulator demonstrates that MaskFuser significantly enhances the performance of existing learning methods, improving both pure imitation learning and safety-centric metrics. Furthermore, MaskFuser's sophisticated fusing model demonstrates improved robustness against impaired sensory inputs, achieved through advanced multi-modality reconstruction learning. These findings suggest that MaskFuser could pave the way for novel feature fusion approaches in future autonomous driving research.

\section*{Acknowledgments}


\bibliographystyle{IEEEtran}
\bibliography{aaai24}

\newpage

\vfill

\end{document}